%% file: main.tex
\documentclass[acmtog]{acmart}

\acmSubmissionID{330}

\usepackage{multirow}
\usepackage{multicol}
\usepackage{array}
\usepackage{enumitem}
\usepackage{multirow}
\usepackage{diagbox}
\usepackage{booktabs} 

\usepackage[ruled]{algorithm2e} 

\SetAlFnt{\small}
\SetAlCapFnt{\small}
\SetAlCapNameFnt{\small}
\SetAlCapHSkip{0pt}

\setcopyright{acmlicensed}
\acmJournal{TOG}
\acmYear{2024} \acmVolume{43} \acmNumber{6} \acmArticle{201} \acmMonth{12}\acmDOI{10.1145/3687903}

\begin{document}
\title{Generative Portrait Shadow Removal}

\author{Jae Shin Yoon}
\email{jaeyoon@adobe.com}
\orcid{0000-0003-0181-4869}
\affiliation{\institution{Adobe Inc.}\country{United States}\city{San Jose}}

\author{Zhixin Shu}
\email{zshu@adobe.com}
\orcid{0009-0004-3878-4805}
\affiliation{\institution{Adobe Inc.}\country{United States}\city{San Jose}}

\author{Mengwei Ren}
\email{mren@adobe.com}
\orcid{0000-0002-7552-1279}
\affiliation{\institution{Adobe Inc.}\country{United States}\city{San Jose}}

\author{Xuaner Zhang}
\email{cezhang@adobe.com}
\orcid{0000-0002-7679-800X}
\affiliation{\institution{Adobe Inc.}\country{United States}\city{San Jose}}

\author{Yannick Hold-Geoffroy}
\email{holdgeof@adobe.com}
\orcid{0000-0002-1060-6941}
\affiliation{\institution{Adobe Research}\country{Canada}\city{Quebec}}

\author{Krishna kumar Singh}
\email{krishsin@adobe.com}
\orcid{0000-0002-8066-6835}
\affiliation{\institution{Adobe Inc.}\country{United States}\city{San Jose}}

\author{He Zhang}
\email{hezhan@adobe.com}
\orcid{0000-0002-7036-6820}
\affiliation{\institution{Adobe Inc.}\country{United States}\city{San Jose}}

\begin{teaserfigure}
\begin{centering}
  \includegraphics[width=0.95\linewidth]{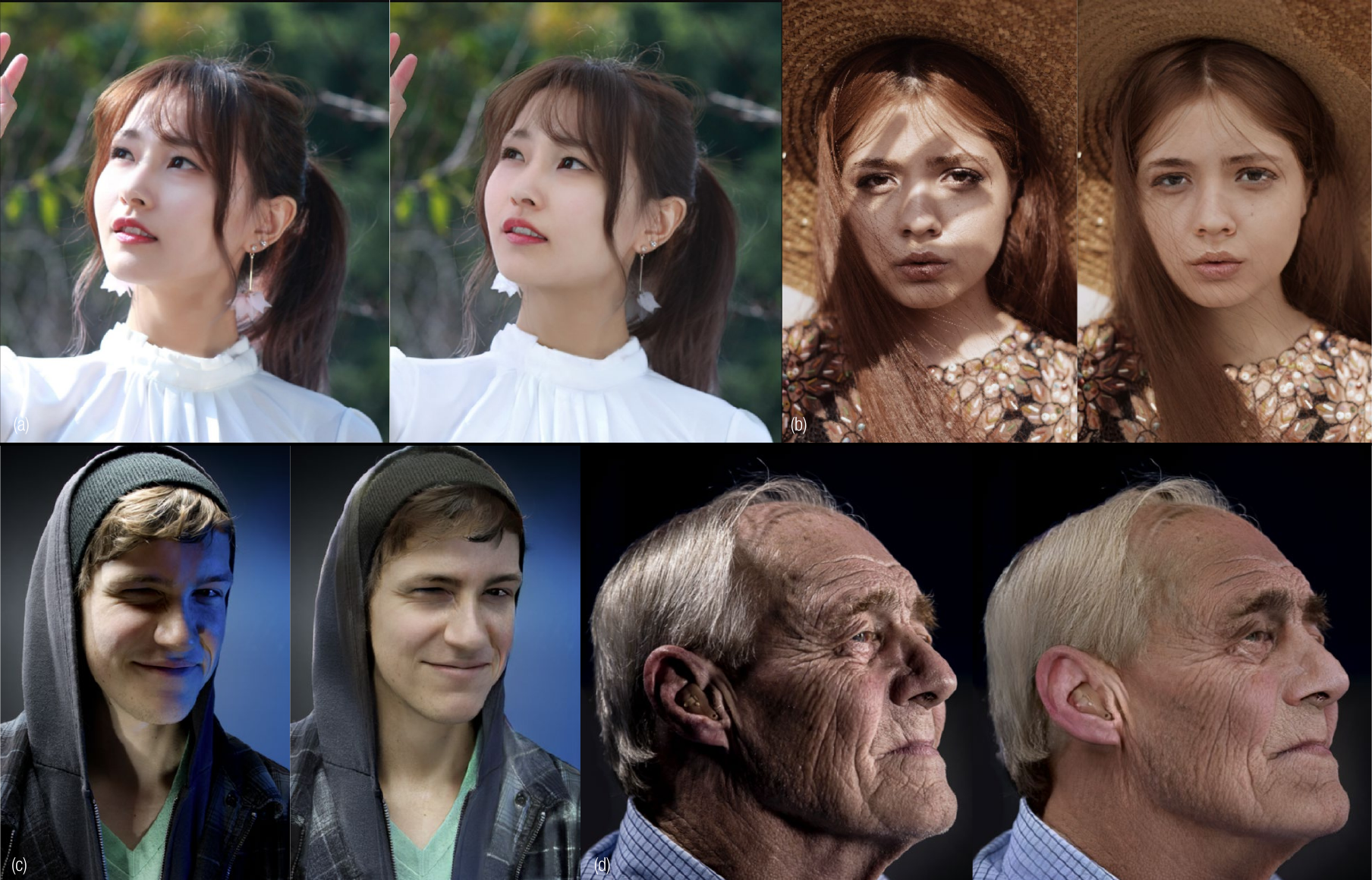}
  \vspace{-2mm}
  \caption{Results from our generative portrait shadow removal method. Our method (a,b) robustly removes the shadows and highlights cast by self and external occlusions; (c) Plausibly estimates natural intrinsic appearance under colored lighting; and (d) effectively preserves the high-frequency details such as skin pores, wrinkles, and hair. }
  \label{fig:teaser}
  \end{centering}
\end{teaserfigure}

\begin{abstract}
\input{sec/0_abstract}

\end{abstract}

%
%



\bibliographystyle{ACM-Reference-Format}

\maketitle
\input{sec/1_intro}

\input{sec/2_related}
\input{sec/3_method}

\input{sec/4_experiments}

\input{sec/5_conclusion}

\begin{figure}
  \centering
    \includegraphics[width=1\columnwidth]{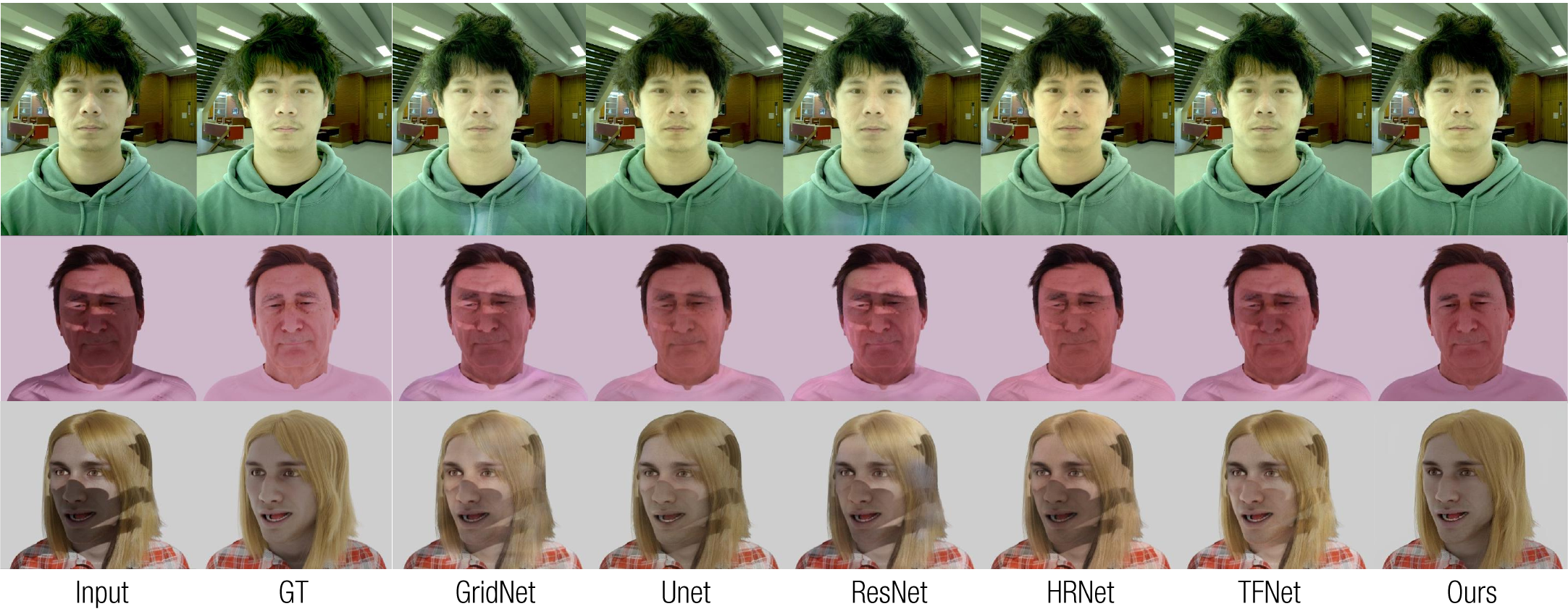}
  \vspace{-0.7cm}
  \caption{\textbf{Comparison} with other baselines on the validation data. 
  }
  \label{Fig:validation}
\end{figure}

\begin{figure}[t]
  \centering
    \includegraphics[width=1\columnwidth]{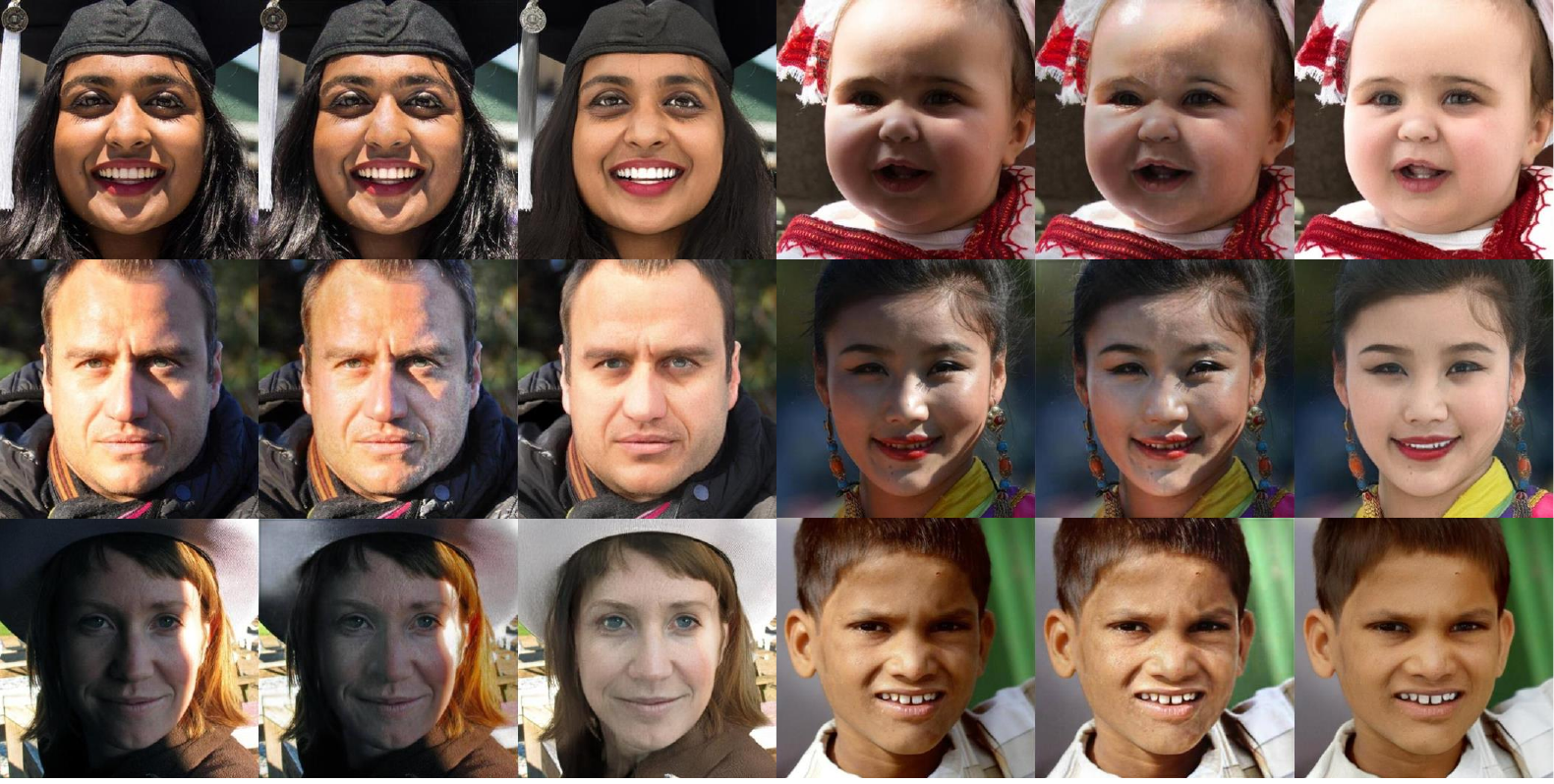}
  \vspace{-0.7cm}
  \caption{\textbf{Comparison.} Left: input, Middle: \cite{han2021deep}, Right: Ours.
  }
  \label{Fig:com_dpl}
\end{figure}

\begin{figure*}
    \centering
    \includegraphics[width=0.85\linewidth]{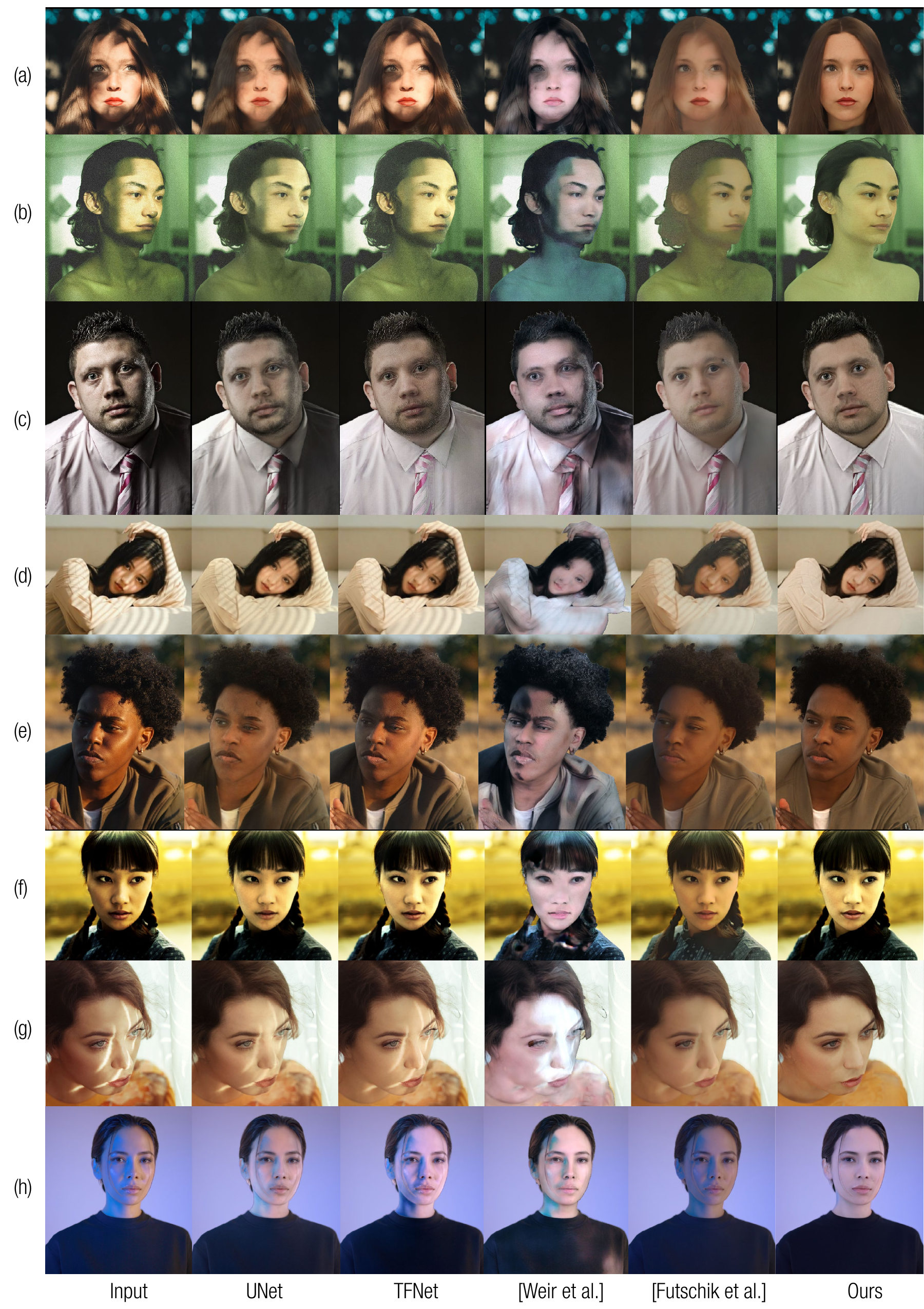}
    \vspace{-3mm}
    \caption{\textbf{Comparison} with UNet, TransformerNet (TFNet), Deep Portrait Delighting \cite{weir2022deep} and Controllable Light Diffusion \cite{futschik2023controllable}.}
    \label{fig:qual}
\end{figure*}




\begin{figure}[t]
  \centering
    \includegraphics[width=1\columnwidth]{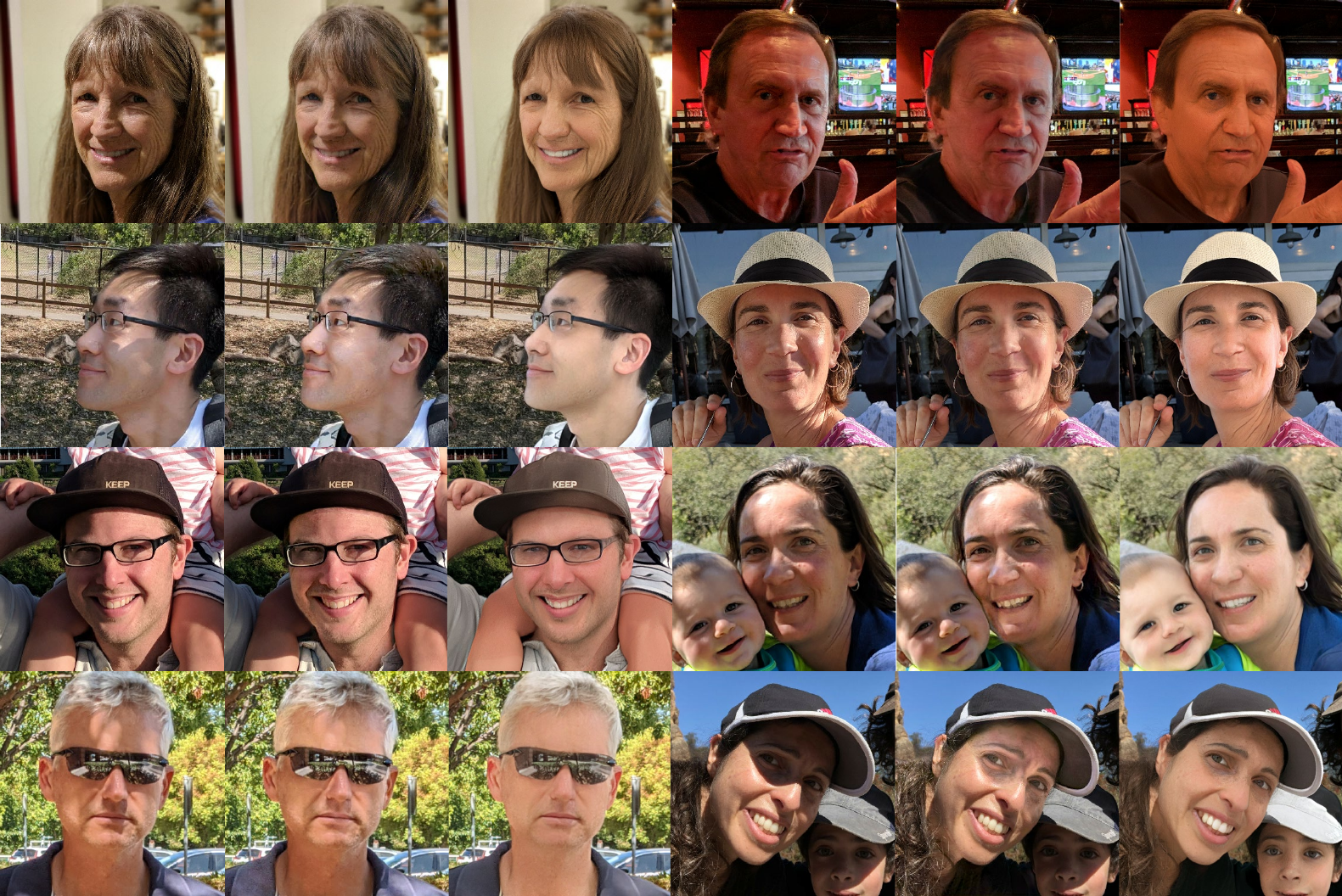}
  \vspace{-0.7cm}
  \caption{\textbf{Comparison.} Left: input, Middle: \cite{zhang2020portrait}, Right: Ours. 
  }
  \label{Fig:comp_cecilia}
    \centering
    \includegraphics[width=1\columnwidth]{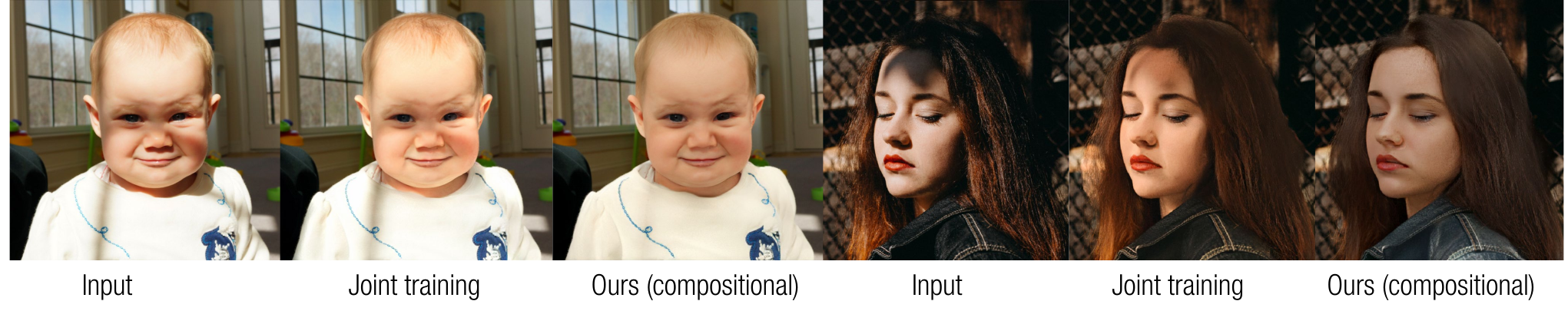}
  \vspace{-0.7cm}
  \caption{\textbf{Ablation study}: Joint training with background harmonization and shadow removal, compared to our compositional repurposing framework. 
  }
  \label{abl:JR}
    \centering
    \includegraphics[width=1\columnwidth]{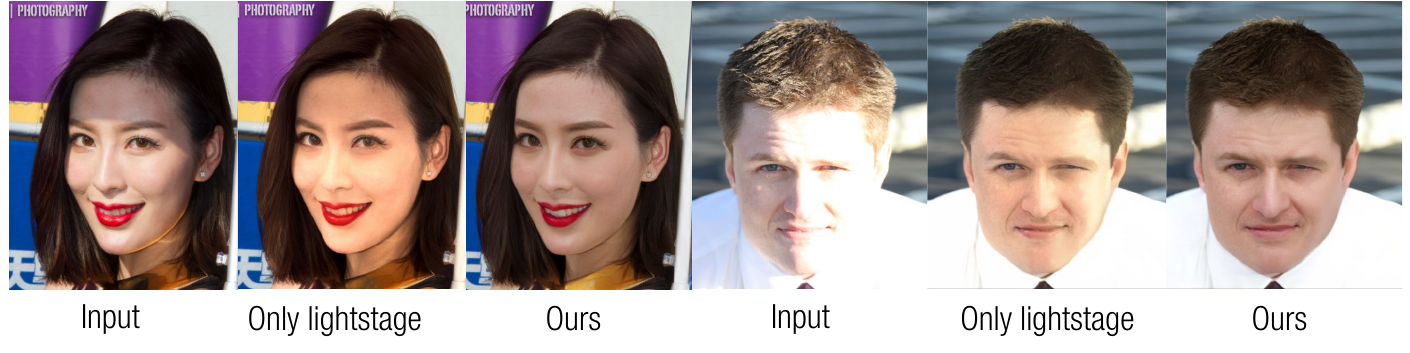}
  \vspace{-0.7cm}
  \caption{\textbf{Data ablation study}: Training only with lightstage data. 
  }
  \label{abl:no_synth}
  \centering
    \includegraphics[width=1\columnwidth]{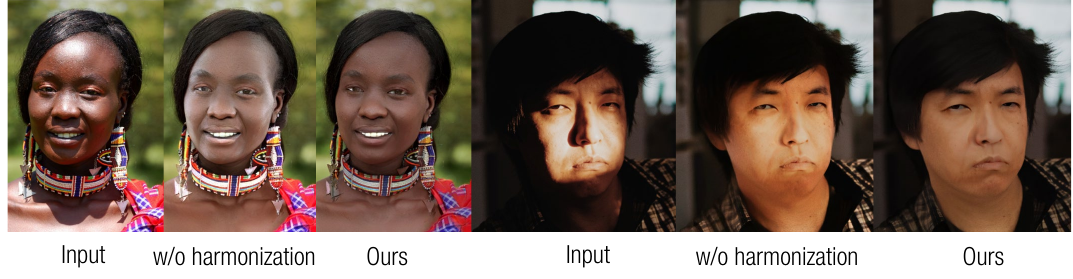}
  \vspace{-8mm}
  \caption{\textbf{Ablation study}: Repurposing without background harmonization. 
  }
  \label{abl:no_harmo}
  
    \centering
    \includegraphics[width=1\columnwidth]{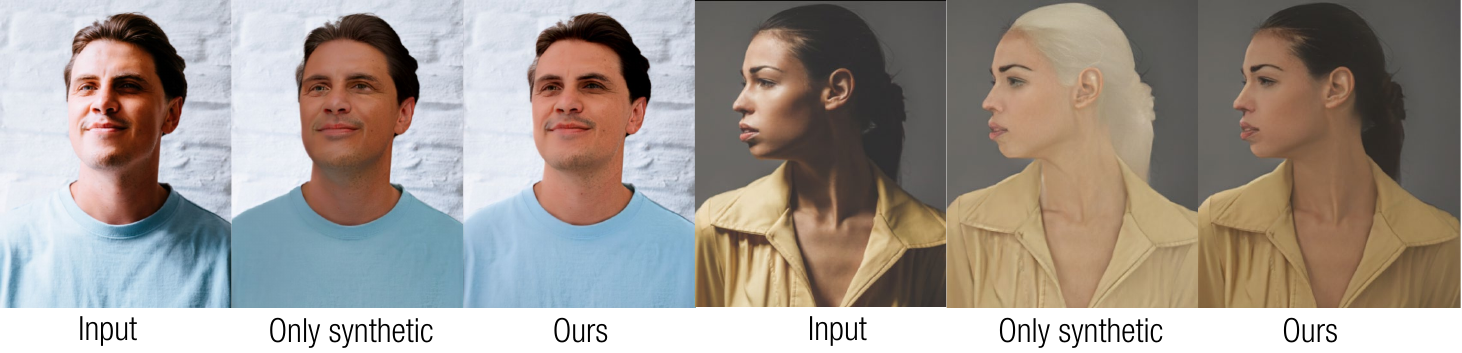}
  \vspace{-0.8cm}
  \caption{\textbf{Data ablation study}: Training only with synthetic humans.
  }
  \label{abl:no_ligghtstage}
\end{figure}





\begin{figure}

    \centering
    \includegraphics[width=1\columnwidth]{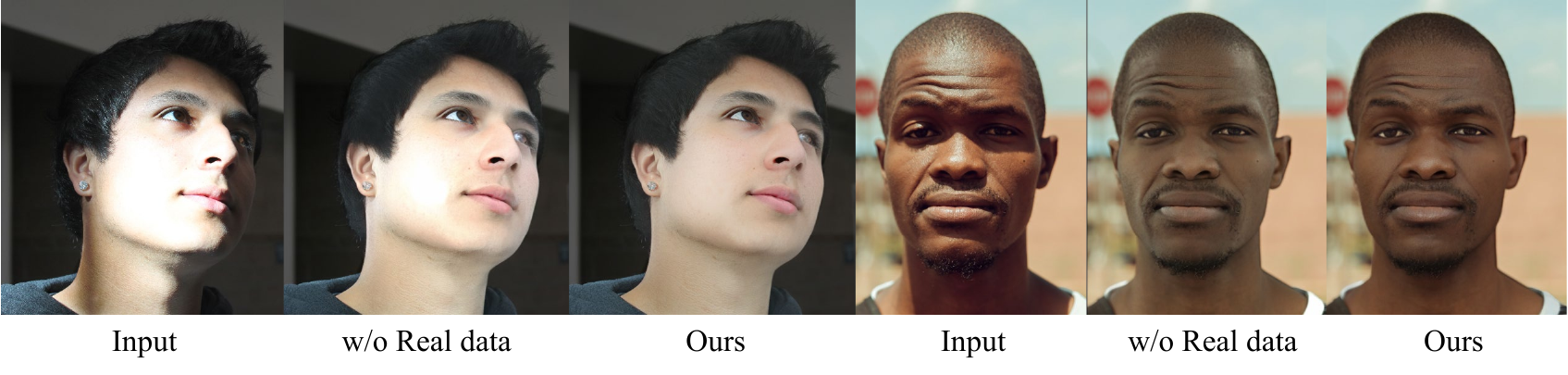}
  \vspace{-0.9cm}
  \caption{\textbf{Data ablation study}: Training without many real-world data.
  }
  \label{abl:no_real}

   \centering
    \includegraphics[width=1\columnwidth]{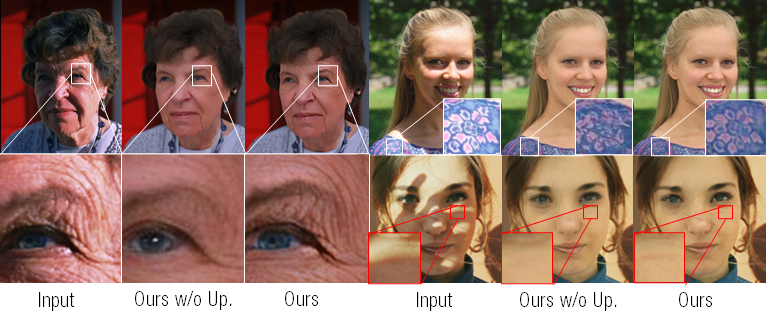}
  \vspace{-0.9cm}
  \caption{\textbf{Ablation study} on our guided upsampling module.
  }
  \label{abl:up}

    \centering
    \includegraphics[width=1\columnwidth]{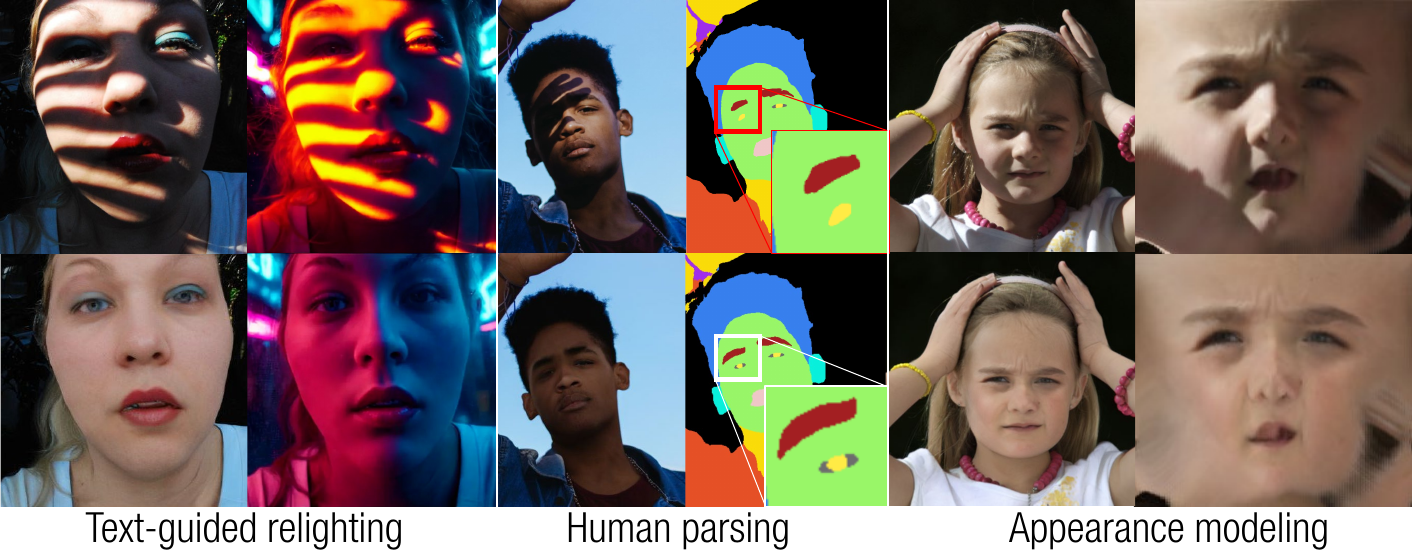}
  \vspace{-0.7cm}
  \caption{\textbf{Application}: Our shadow removal results (bottom) from the inputs (top) improve the quality of downstream tasks.
  }
  \label{application1}

    \centering
    \includegraphics[width=1\columnwidth]{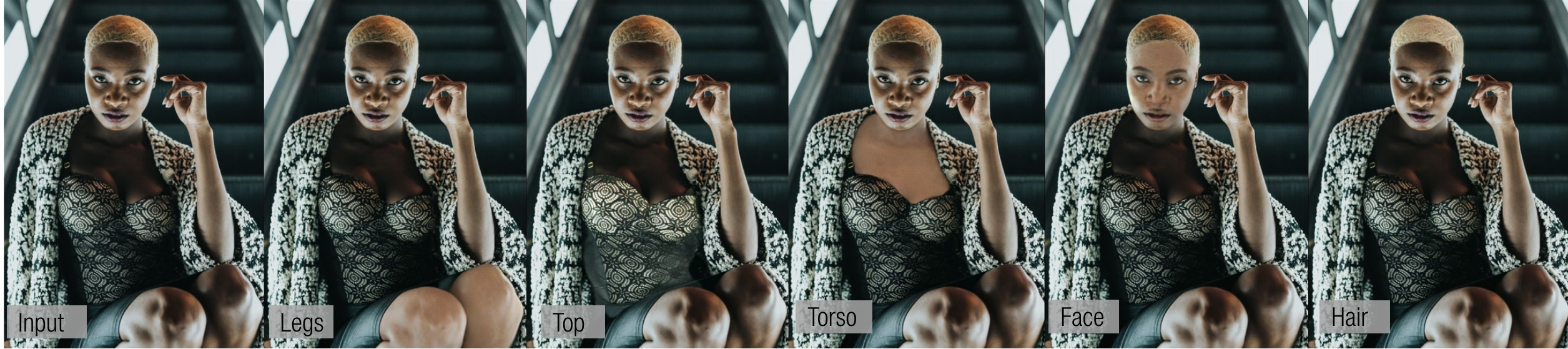}
  \vspace{-0.7cm}
  \caption{\textbf{Application}: Part-aware shadow removal and editing.
  }
  \label{application2}

      \centering
    \includegraphics[width=1\linewidth]{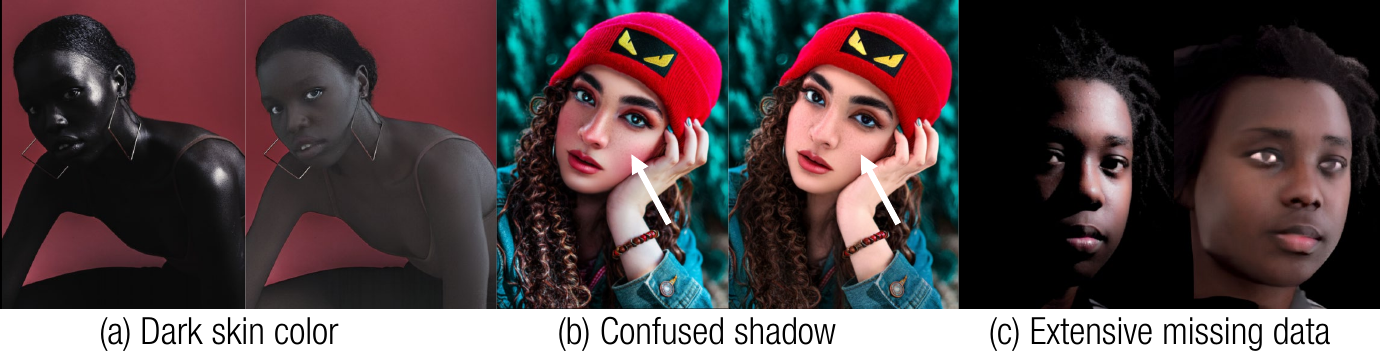}
    \vspace{-6mm}
    \caption{\textbf{Limitation} of our generative shadow removal model.}
    \label{fig:limitation}
\end{figure}






\bibliography{bibtex}
\renewcommand\floatpagefraction{.001}

\input{sec/6_supple}

\end{document}

%% file: sec/0_abstract.tex
We introduce a high-fidelity portrait shadow removal model that can effectively enhance the image of a portrait by predicting its appearance under disturbing shadows and highlights.
Portrait shadow removal is a highly ill-posed problem where multiple plausible solutions can be found based on a single image. For example, disentangling complex environmental lighting from original skin color is a non-trivial problem. 
While existing works have solved this problem by predicting the appearance residuals that can propagate local shadow distribution, such methods are often incomplete and lead to unnatural predictions, especially for portraits with hard shadows.
We overcome the limitations of existing local propagation methods by formulating the removal problem as a generation task where a diffusion model learns to globally rebuild the human appearance from scratch as a condition of an input portrait image. 
For robust and natural shadow removal, we propose to train the diffusion model with a compositional repurposing framework: 
a pre-trained text-guided image generation model is first fine-tuned to harmonize the lighting and color of the foreground with a background scene by using a background harmonization dataset; and then the model is further fine-tuned to generate a shadow-free portrait image via a shadow-paired dataset.
To overcome the limitation of losing fine details in the latent diffusion model, we propose a guided-upsampling network to restore the original high-frequency details (\textit{e.g.,} wrinkles and dots) from the input image.
To enable our compositional training framework, we construct a high-fidelity and large-scale dataset using a lightstage capturing system and synthetic graphics simulation.
Our generative framework effectively removes shadows caused by both self and external occlusions while maintaining original lighting distribution and high-frequency details. Our method also demonstrates robustness to diverse subjects captured in real environments.


%% file: sec/1_intro.tex
\section{INTRODUCTION}
Photos of a portrait scene capture a unique moment. While we later find that they include many disturbing shadows, physically restoring such past moments for recapturing is often impossible.
In this paper, we introduce a high-fidelity model that can preserve the moment of the photo at a specific scene and its lighting condition while removing the disturbing shadows and highlights on the portraits as shown in Figure~\ref{fig:teaser}.  
Our foundational model can expedite many tedious processes of existing image editing tools (\textit{e.g.,} Gimp, FaceApp, Photoshop) for portrait shadow removal as well as be used as post-processing for many applications such as portrait relighting and modeling of a personalized avatar with clean appearance.

\begin{figure}[t]
  \centering
    \includegraphics[width=1\columnwidth]{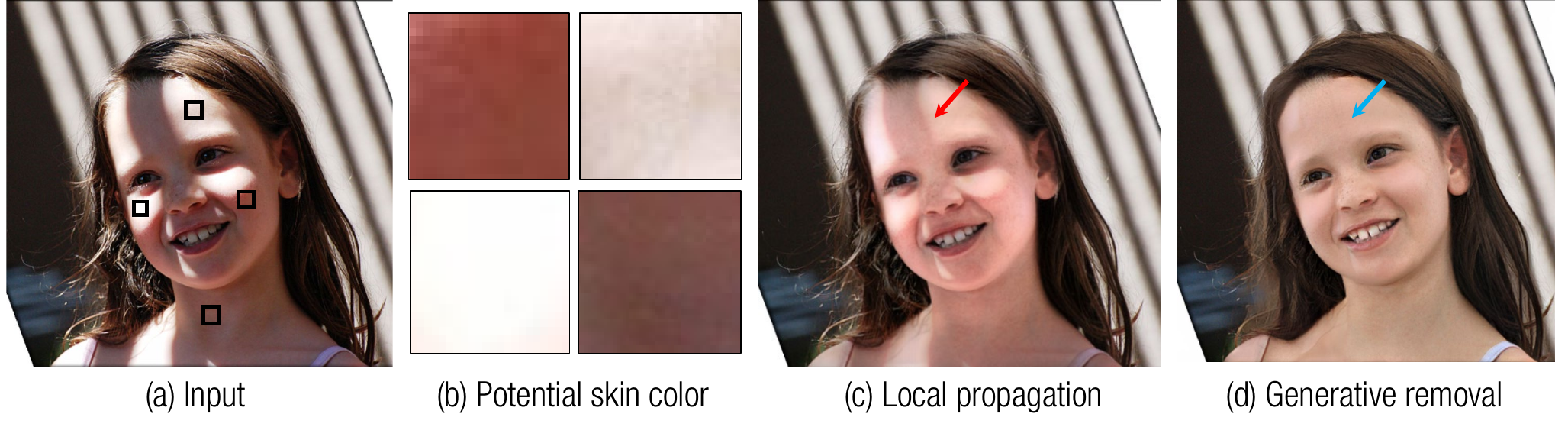}
  \vspace{-0.7cm}
  \caption{\textbf{Motivation.} (a) Input image of a portrait under external shadow. (b) Appearance samples on the skin where any of them can be a plausible solution. (c) The results with a residual appearance prediction model (\textit{e.g., } UNet \cite{ronneberger2015u}) which is designed to propagate the local residual appearance. This is often trapped by the local minima (\textit{e.g.,} strong shadow boundary). (d) We generatively remove the shadow by globally re-building the shadow-free portrait images from scratch.
  }
  \label{Fig:motivation}
\end{figure}

Removing disturbing shadows from a portrait scene is challenging due to the significant ill-posedness inherent in human appearance modeling. 
For example, in Figure~\ref{Fig:motivation}(b), what would be the accurate skin color of the subject beneath the shadow or highlight? 
Any of the four samples can be a potential solution, which reveals the nature of one-to-many mapping problems. This issue is further exacerbated in real-world scenes, under dynamic lighting conditions.

To overcome this challenge, many existing works~\cite{zhang2020portrait,weir2022deep,weir2023lighting,futschik2023controllable} have approached the shadow removal task in a local propagation way: A model predicts the residual appearance that can diffuse the neighboring pixels. 
However, these methods are often trapped by local minima, which produces suboptimal results, particularly around the strong boundary of the shadow, as shown in the third column of Figure~\ref{Fig:motivation}.  
Further, some of such local propagation methods that learn inverse rendering datasets (\textit{e.g.,} intrinsic albedo decomposition~\cite{mei2024holo,hou2022face,mei2023lightpainter,nestmeyer2020learning})
often drastically change the lighting distribution of the portrait. This produces significant lighting inconsistency between the predicted foreground and the original background scene whose composition is highly unnatural.

We address those limitations by casting the portrait shadow removal problem as a generation task for shadow-free portrait images.
A generative diffusion model learns to reconstruct the image of a portrait from scratch as a condition of noisy data (\textit{i.e.,} portraits under disturbing shadows), which produces globally coherent shadow-free images.
To effectively keep the original lighting distribution and perform robust shadow removal, we train our framework with compositional repurposing of existing image priors: a pre-trained text-guided image generation model (\textit{e.g.,} latent diffusion~\cite{rombach2022high}) is first optimized to harmonize the color and lighting between a portrait image and a different background scene. It is further fine-tuned to generate a shadow-free portrait appearance under the same lighting conditions as the background scene.  


However, the current state-of-the-art text2image diffusion models \cite{rombach2022high,sora} are mainly working in latent space, which may result in the loss of identity or details from the input portrait.  Hence, we propose a refinement module to help restore the input's high-frequency details which are often smoothed out in the denoising process of the diffusion model. To this end, we develop a guided upsampling network that learns to extract the high-frequency details from the input image and the low-frequency lighting distribution from the generated shadow-free image. 

To model background harmonization as well as natural and robust shadow removal, we construct a high-quality shadow removal dataset using 1. Data captured and rendered by a lightstage, 2. Synthetically rendered humans, and 3. Augmented Real-world portraits leveraging 3D geometry such as depth and normals. For the lightstage dataset, portrait images under diverse lighting and background scenes are rendered using OLAT (One-Light-at-A-Time) images lit by augmented environment maps. A portrait image with mostly uniform lighting is also rendered using a diffused environment map. A synthetic human model with metric-level geometry is rendered with external shadows by placing random objects between the light and the head, which simulates diverse novel shadows. Such novel shadow simulation is also applied to many real-world portraits with estimated geometry data (\textit{e.g.,} estimated depth and normal). 


In the experiments, our model demonstrates natural shadow removal results robust to self- and external occlusions while ensuring 1) its original lighting distribution minimally deviates from the background scene; and 2) the person's identity and its high-frequency details are well preserved.
The compositional modeling of portrait harmonization and shadow removal allows our diffusion model to effectively reveal the appearance even under colored lighting and shadow, outperforming existing local appearance propagation methods.
%
%
Our shadow removal model enables applications such as high-fidelity portrait relighting, human parsing with increased robustness, and clean human appearance modeling.

In summary, our main contributions include:
\begin{itemize}
    \item The formulation of generative portrait shadow removal with our compositional repurposing framework.
    \item A data construction pipeline that can push the visual distribution of four data points: lighting, shadow by self-occlusion, shadow by external occlusion, and identity. 
    \item A complete shadow removal pipeline with a guided-upsampling network that preserves the portrait identity with minimum loss of high-frequency details. 
    \item Extensive experiments that include comparisons, ablation studies, and applications.
\end{itemize}

%% file: sec/2_related.tex
\section{RELATED WORK}

\subsection{Portrait Shadow Removal}
Existing methods have solved the portrait shadow removal problem by introducing variational forms of local appearance propagation.
\cite{zhang2020portrait} introduced a method to synthesize realistic shadows on real-world portraits from which a residual network learns jointly with lightstage data to enhance the portrait shadow condition with symmetry priors. 
Deep Portrait Delighting~\cite{weir2022deep} proposed a Unet-based residual network that can predict the soft shadow with shading offset regularization to separate the shadow from texture.  
\cite{weir2023lighting} further introduced a de-lighting model that learns a set of human albedo (\textit{i.e.,} an intrinsic appearance) data to infer the shadow-free images with effective data augmentation (\textit{e.g.,} color shifting).
{\cite{he21unsupervised} introduced an unsupervised approach to learn a shadow removal model by utilizing a generative prior of a shadow-free face images.}
Blind Removal~\cite{liu2022blind}) particularly addressed the removal problem of the outside shadow on the face region by learning from simulation data of gray-scale images.
\cite{futschik2023controllable} introduced an image-based method that can diffuse the light by learning from panorama environment maps with sub-surface scattering to suppress the strong local shadow on a portrait.
While the above methods have shown promising results robust to a soft shadow, their shadow removal results on portraits with strong casting shadows are often incomplete and yield artifacts such as blurs.

\subsection{Human Relighting}
Existing image-based human relighting methods, in general, include an inverse rendering pipeline such as intrinsic decomposition where the albedo is often used as a proxy of shadow-free image.
The Unet-based network is most widely used by many relighting works~\cite{mei2024holo,hou2022face,mei2023lightpainter,nestmeyer2020learning,pandey2021total,kim2024switchlight,hou2024compose} to predict the shadow-free albedo appearance.
To effectively suppress the disturbing shadows, some works have adopted more advanced neural networks such as  HRNet~\cite{ji2022geometry}, ReidualNet~\cite{yeh2022learning,lagunas2021single,tajima2021relighting} or utilized a video for per-subject UV optimization for shadow-free appearance~\cite{wang2023sunstage}. On top of the predicted albedo, novel lighting is added under the control of environment map conditions~\cite{yeh2022learning,mei2024holo,pandey2021total,kim2024switchlight,ji2022geometry}, point and directional lighting with ray tracing~\cite{hou2022face,nestmeyer2020learning}, user-specified scribbles~\cite{mei2023lightpainter}, physically based rendering~\cite{wang2023sunstage}, spherical harmonics~\cite{lagunas2021single,tajima2021relighting}.
{Computational 3D models, \textit{e.g.,} 3D face model with the combination of 3D pose estimation~\cite{ranjan2023facelit}
, have been often used as a prior to learn the relighting function without labeled data.}
While the portrait relighting results were promising, their albedo prediction results are often too flat (due to the nature of intrinsic appearance); and their lighting distribution largely changes from the one in the original image.

Some works have performed the relighting without inverse rendering techniques but with direct manners. Various lighting variables such as environment maps~\cite{sun2019single}, preset reference portrait image~\cite{song2021half}, low-resolution background image~\cite{ren2023relightful}, coarse shading estimate~\cite{ponglertnapakorn2023difareli}, or spherical harmonics~\cite{hou2021towards,zhou2019deep} are conditioned onto the latent space of a network which directly decodes relit images. 
Since such direct approaches are designed to change the shadow direction instead of its removal, their results are often sub-optimal as shadow-free portrait images.

\begin{figure*}[t]
  \centering
    \includegraphics[width=2\columnwidth]{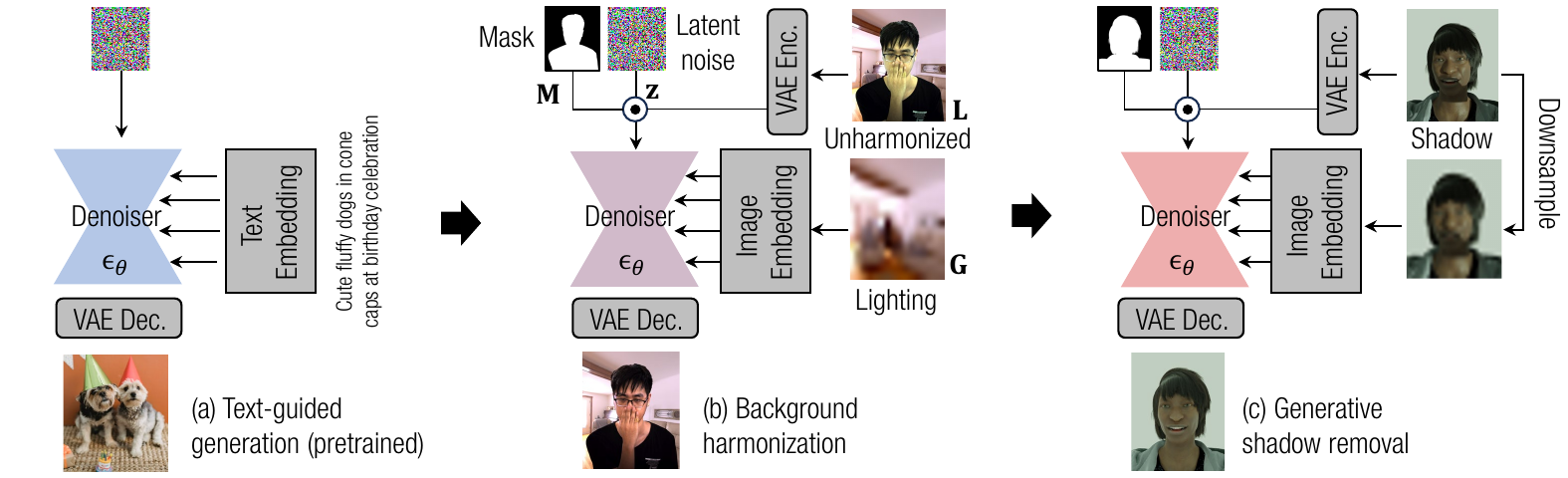}
  \caption{\textbf{Overview of our compositional repurposing framework.} (a) A denoising diffusion model learns massive images and text pairs to generate an image from a text, which forms a large image prior. We perform a series of repurposing of this prior: (b) The diffusion model learns to generate the harmonized portrait images with respect to the background scenes as a condition of a downsampled background lighting map. (c) The diffusion model is further fine-tuned to generate a shadow-free portrait where the downsampled inputs are used as the lighting map. Here, Enc. and Dec. mean Encoder and Decoder, respectively. 
  }
  \label{Fig:overview}
\end{figure*}


\subsection{General Scene Intrinsic Decomposition}
The image of a portrait is a special case of a general scene, and therefore, applying a general-scene intrinsic decomposition method to the portrait scene is possible.
Optimization-based approaches have been using various objective cues and priors such as smooth shading~\cite{chen2013simple}, grayscale shading \cite{garces2012intrinsic, grosse2009ground, zhao2012closed}, albedo sparsity \cite{shen2011intrinsic, bell2014intrinsic, meka2021real, bi20151}, 3D geometry \cite{hachama2015intrinsic, wu2014real, yu2013shading, zollhofer2015shading},
user interaction \cite{meka2017live, bousseau2009user}, multi-view images~\cite{ye2023intrinsicnerf}.
{Learning-based approaches~\cite{fan2018revisiting, liu2020unsupervised, li2018cgintrinsics,jin2022reflectguidance,kim2016unified, zhou2019glosh, luo2020niid,luo2023crefnet,sfsnetSengupta18} aim to learn high-quality labeled data for albedo and other intrinsic values such as shading, normal, reflectance, such that, it expedites the reconstruction of the accurate intrinsic values.}
While they produce promising general scene albedo results, the quality is often significantly degraded when they are applied to a portrait scene due to the mismatch of visual statistics between portrait and general scenes. 

\subsection{General Scene Shadow Removal}
Existing general scene shadow removal models aim to remove the shadow cast on the ground or wall formed by external occluders. As a general framework, they first detect the shadow and then remove the shadow conditioned by the detection using hand-crafted shadow anchor point modeling~\cite{arbel2010shadow}, vanilla convnet~\cite{khan2015automatic}, stacked neural networks~\cite{wang2018stacked}, direction aware attention map~\cite{hu2019direction}, progressive residual process~\cite{ding2019argan}, multi-scale image decomposition~\cite{zhang2019effective}, and context embedding framework~\cite{chen2021canet}.
To improve the efficiency of the shadow removal framework, some works~\cite{liu2023shadow,kubiak2024s3r,qu2017deshadownet} introduced an end-to-end pipeline that can directly output shadow-free images from a single input image using residual networks~\cite{liu2023shadow}, cycle consistency~\cite{kubiak2024s3r}, compositional context embedding~\cite{qu2017deshadownet}, and harmonization~\cite{ma2016appearance}. 
However, applying those models to portraits does not work well due to significant inconsistency in the visual distribution between general pictures and portraits.

%% file: sec/3_method.tex
\section{Methodology}
\subsection{Preliminaries}
We train a diffusion model~\cite{ho2020ddpm} to produce an image through the process of denoising a noise map. The training procedure involves both a forward and a backward step.
In the forward step, it constructs intermediate noisy images by gradually adding Gaussian noise to the noise-free data under a Markovian chain:
\begin{align}
    \mathbf{x}_t = \sqrt{\overline{\alpha}_t} \mathbf{x}_0 + \sqrt{1 - \overline{\alpha}_t} \boldsymbol\epsilon \text{,}
\end{align}
where $\boldsymbol\epsilon \!\sim\! \mathcal{N}(\mathbf{0}, \mathbf{I})$ is the Gaussian noise, $\mathbf{x}_0$ is a clean image, $\mathbf{x}_t$ is the noisy image at time step $t$, and $\overline{\alpha}_t$ is computed from a fixed variance schedule.
This forward process can be extended to latent images~\cite{rombach2022high}:
\begin{align}
    \mathbf{z}_t = \sqrt{\overline{\alpha}_t} \mathbf{z}_0 + \sqrt{1 - \overline{\alpha}_t} \boldsymbol\epsilon \text{,}
\end{align}
where $\mathbf{z}_0$ is the latent features extracted by a pre-trained image encoder network (\textit{e.g.,} Variational Autoencoder (VAE)~\cite{pinheiro2021variational}), and $\mathbf{z}_t$ is the noisy latent features at time $t$.

In the backward process, a denoiser such as \textit{e.g.,} U-Net~\cite{ronneberger2015u} \cite{ronneberger2015u} is trained to construct a clean image by generating the noise at a time step $t$ with the following objectives:
\begin{align}
    \mathcal{L} = \mathbb{E}_{\mathbf{z}_0, \boldsymbol\epsilon \sim \mathcal{N}, t}
    \!\left[ \|\boldsymbol\epsilon - \boldsymbol\epsilon_{\boldsymbol\theta}(\mathbf{z}_t, t)\|^2_2 \right] \text{,}
    \label{eq:obj}
\end{align}
where $\boldsymbol\epsilon_{\boldsymbol\theta}(\cdot)$ is the noise prediction function. 

There are two ways to control the local and global properties of the image generation from $\boldsymbol\epsilon_{\boldsymbol\theta}(\cdot)$. 

\noindent 1) Local control:  Similar to existing conditional diffusion frameworks (\textit{e.g.,} InstructPix2Pix~\cite{brooks2023instructpix2pix}), a spatially aligned conditional map can contribute its local information such as edges and pose map to the generated images by concatenating it with the latent noise $\mathbf{z}_t$ \textit{i.e.,} $\mathbf{z}_t\rightarrow\{\mathbf{z}_t, \mathbf{L}\}$ where $\mathbf{L}\in\mathbb{R}^{W\times H\times N}$. Since they are spatially aligned, the denoiser can effectively borrow some local information from $\mathbf{L}$ to replace such properties in outputs.
This local condition can be extended to the latent space by utilizing VAE encoder where existing works~\cite{brooks2023instructpix2pix} have demonstrated that conditioning the local variables in the same latent space as the one of the latent noise is effective to guide the local properties of the image generation, \textit{i.e.,} $\{\mathbf{z}_t, \mathbf{L}\}\rightarrow\{\mathbf{z}_t, \mathbf{z}_{\mathbf{L}})\}$ where $\mathbf{z}_{\mathbf{L}}$ is the time-invariant latent images encoded by VAE. 

\noindent 2) Global control: Global properties (\textit{e.g.,} semantics, text, and lighting) of a scene can be conditioned on the denoiser $\boldsymbol\epsilon_{\boldsymbol\theta}$ in an embedding space of a global conditional variable $\mathbf{G}$ using existing subspace embedding modules such as CLIP~\cite{radford2021learning} for text or DINO~\cite{caron2021emerging} for image. Unlike local control, the global variables are not spatially aligned, and therefore, they are often conditioned via the attention mechanism (\textit{e.g.,} cross-attention~\cite{vaswani2017attention}) to allow the denoiser to find the correspondences between its intermediate features and global conditioning. 


The backward denoising process can consider these local and global control signals by minimizing the following objectives:
\begin{align}
    \mathcal{L} = \mathbb{E}_{\mathbf{z}_0, \mathbf{y}, \boldsymbol\epsilon \sim \mathcal{N}, t} \! \left[ \| \boldsymbol\epsilon - \boldsymbol\epsilon_{\boldsymbol\theta}(\{\mathbf{z}_t, \mathbf{L}\}, t, \boldsymbol\tau(\mathbf{G})) \|^2_2 \right],
    \label{eq:obj_cond}
\end{align}
where $\boldsymbol{\tau}(\cdot)$ is the subspace embedding function that projects the global control variable to the latent space.
We learn this objective in a compositional way to develop a foundational generative model for portrait shadow removal as described in the sections below.

\subsection{Compositional Repurposing}
Given a pre-trained text-to-image generation model that learns from large-scale visual data with Equation~\ref{eq:obj}, we perform a series of repurposing to accomplish two tasks: 1) light-aware background harmonization and 2) shadow-free image generation. The overview of our training pipeline is described in Figure~\ref{Fig:overview}.

\subsubsection{Light-Aware Background Harmonization}
The pre-trained text-guided image generation model is fine-tuned to predict the noise that generates the clean portrait image which harmonizes with a background scene, by optimizing the following objectives:
\begin{align}
    \mathcal{L} = \mathbb{E}_{\mathbf{z}_0, \mathbf{y}, \boldsymbol\epsilon \sim \mathcal{N}, t} \! \left[ \| \boldsymbol\epsilon - \boldsymbol\epsilon_{\boldsymbol\theta}(\{\mathbf{z}_t, \mathbf{z}_{L},\mathbf{M}\}, t, \boldsymbol\tau(\mathbf{G})) \|^2_2 \right],
    \label{eq:obj_harmon}
\end{align}
where $\textbf{M}\in\mathbb{R}^{W\times H\times N}$ is the downsampled foreground mask, which is directly concatenated with the noisy latent image $\mathbf{z}_{t}$ to guide the attention of the foreground region during the denoising process. 
%
$\mathbf{z}_{L}$ is the time-invariant conditional latent features projected from the input unharmonized image $\mathbf{L}$ using VAE encoder. $\mathbf{z}_{L}$ thus shares a common latent space with $\mathbf{z}_{t}$. 
$\mathbf{G}$ is the background image that guides the global illumination in the embedding space projected from the image embedding module $\boldsymbol\tau(\cdot)$, \textit{i.e.,}~DINOv2~\cite{oquab2023dinov2}.
%
Following ~\cite{ren2023relightful}, the un-harmonized image $\mathbf{L}$ is created by composing the original foreground image with a novel background. We use the downsampled background image of $\mathbf{G}$ as a lighting map.
To support the different channel number of the denoiser $\boldsymbol\epsilon_{\boldsymbol\theta}$ from the pretrained text-guided image generation model,
we change the first layer of the network that matches the input modality for background harmonization.
The clean latent space $\mathbf{z}_{0}$ is constructed by projecting the ground-truth harmonized data (captured from lightstage as described in the data Section~\ref{sec:data}) using a VAE encoder.

\subsubsection{Generative Portrait Shadow Removal}
The background harmonization model is further fine-tuned to predict the noise that generates the clean image of a portrait image which includes minimum disturbing shadow and highlights.
We perform this repurposing by minimizing the same objectives in Equation~\ref{eq:obj_harmon} while switching the local and global conditional variables:
The input portrait image with shadows and highlights is used to construct time-invariant local conditional features $\mathbf{z}_{L}$, and the shadow-free portrait image is used to construct the ground-truth latent features $\mathbf{z}_{0}$. 
For the global conditional variable $\mathbf{G}$, we use the downsampled image from the input portrait image $\mathbf{L}$ as a lighting image. 
During the repurposing for the shadow removal, we use a smaller learning rate than the one used for harmonization to minimize the Catastrophic forgetting~\cite{mccloskey1989catastrophic} underlying the sequential learning problem.
In inference time, the diffusion model generates the shadow-free portrait images that are well-harmonized with background scenes by effectively preserving the original lighting distribution from the input image.
While learning with background harmonization and shadow removal together is an alternative, we experimentally found that such joint training makes the model fall into the suboptimal point whose shadow removal results are largely degraded.

\begin{figure}[t]
  \centering
    \includegraphics[width=1\columnwidth]{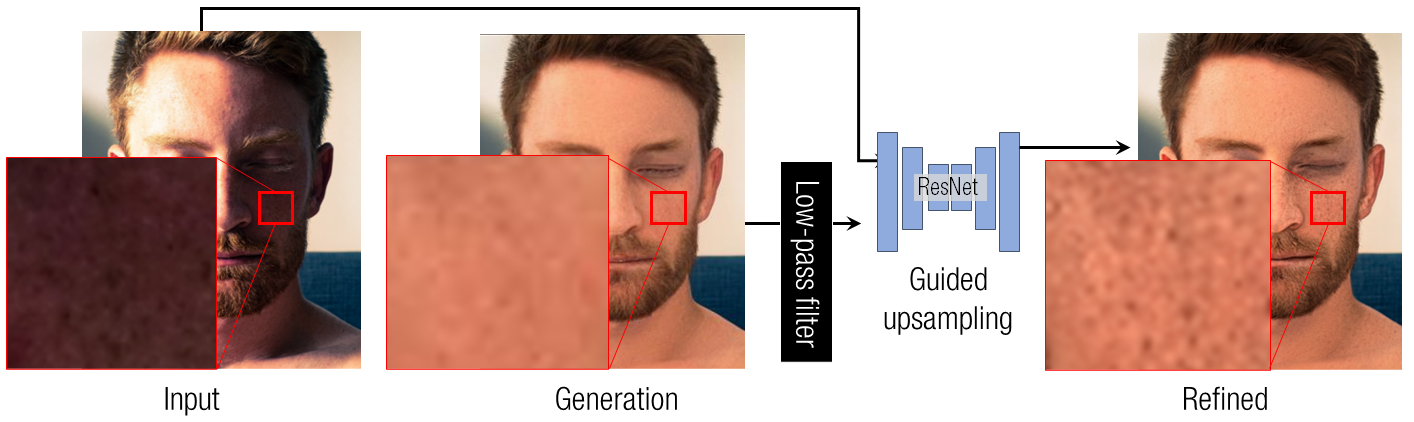}
  \caption{Guided upsampling module that combines the low-frequency components of the generation and high-frequency details from the original input.
  }
  \label{Fig:guided_refine}
\end{figure}

\subsection{Guided Refinement}
Due to the nature of the denoising process of a generative diffusion model, the loss of high-frequency details, \textit{e.g.,} pore, wrinkles, and clothing patterns, is often unavoidable.
%
Therefore, as a post-processing at inference time, we apply a lightweight guided upsampling module that can restore the original details of the portrait while keeping the predicted shadow distribution:
\begin{align}
    \mathcal{I}_{\rm refined} = f(\mathbf{I}_{\rm input},  l(\mathbf{I}_{\rm generation}))
    \label{eq:guided_refine}
\end{align}
where $f$ is the upsampling function designed with a small local prediction network, and $l(\cdot)$ is the low pass filter, \textit{e.g.,} Gaussian filter, $\mathbf{I}_{\rm generation}$ is the generated shadow-free image from our diffusion model, and $\mathbf{I}_{\rm input}$ is the input image with shadows. Inspired by existing studies~\cite{gryka2015learning,khan2015automatic,sanin2012shadow}, our key assumption is that shadows are typically associated with low-frequency components of an image to represent overall lighting distribution, and a network can learn to combine the low-frequency component from the shadow-free image and high-frequency details from the original input image
as shown in Figure~\ref{Fig:guided_refine}

In practice, we use ResidualNet~\cite{he2016deep} and learn from the lightstage data: given a portrait image under a specific lighting condition, we add synthetic disturbance such as blur, noise, and down-sampling similar to existing image restoration works~\cite{zhou2022towards,wang2022survey}. The network learns to predict the original image conditioned on the clean portrait image under different lighting conditions. We train the networks using L2, VGG~\cite{gatys2016image}, and GAN~\cite{isola2017image} losses similar existing works \textit{e.g.,}~\cite{ledig2017photo,chen2023activating,kawar2022denoising}.

\begin{figure}[t]
  \centering
    \includegraphics[width=1\columnwidth]{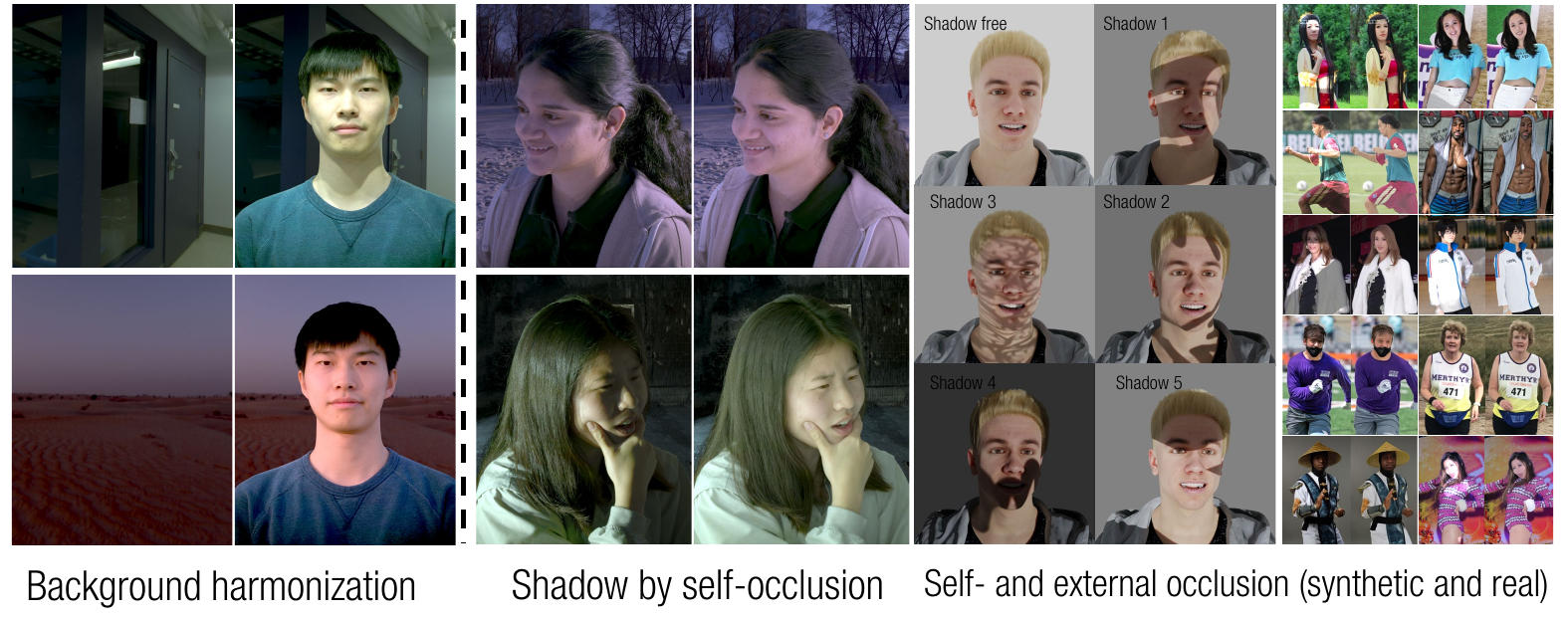}
  \caption{Training dataset for compositional repurposing where the HDR rendering data for background harmonization and portrait shadows by self-occlusion are captured from lightstage, and others are based on graphics simulation.
  }
  \label{Fig:data}
\end{figure}

\subsection{Dataset}~\label{sec:data}
To enable our compositional repurposing pipeline, we collect our dataset from various sources: lightstage system, synthetic humans, and simulation with real data. The data from the lightstage is designed for background harmonization and shadow removal of the person under the self-occluded shadow, and the synthetic and simulated data for both self-occluded and external shadows (\textit{e.g.,} a stark shadow cast by another occluding object) as summarized in Figure~\ref{Fig:data}. {All the training datasets are proprietary data. That is, we internally owned, captured or purchased all the data.}

\subsubsection{OLAT Data Capturing} 
Similar to existing lightstage data capturing methods as discussed in~\cite{debevec2000acquiring}, we collect a set of OLAT (One-Light-at-A-Time) images for 150 unique subjects with varying pose and clothes. Our customized system includes 4 camera views and 160 LED light. We use the MER2-502-79U3M high-speed camera to record the reflectance field of the subject at 5 megapixel resolution and exposure time of 20ms. 


\subsubsection{Background Harmonization}
Similar to \cite{sun2019single}, we relight our OLAT images using diverse HDR environment maps from \cite{hold2019deep,gardner2017learning,polyhaven2023poly} as well as our own HDR bracketed captures using a Ricoh Theta Z1. In particular, we project and tonemap an environment map to obtain a background image and its reference relit portrait. Doing this twice yields pairs of portrait and background images ready for background harmonization, as shown in Figure~\ref{Fig:data}-(left). 

\subsubsection{Diffused Lightstage Dataset}
In addition to the portrait image obtained from the above process, we generate a shadow-free portrait image similar to the method from~\cite{futschik2023controllable}. 
In summary, we render the OLAT portraits again with an energy-preserved blurred version of the environment map. This minimizes self-occluded shadows and diffuses the lighting, while keeping global lighting such as ambient occlusion, as seen in Figure~\ref{Fig:data}. 

\subsubsection{Shadow Simulation with Synthetic Humans} We use a few hundred synthetic humans and render the shadow using point-light-based ray tracing. 
For example, given a synthetic 3D portrait model, we randomly place a point light in front of the subject where we also put a random object in between the portrait and lighting so that it simulates occlusion as shown in Figure~\ref{Fig:data}-(third).

\subsubsection{Augmented Real-World Data}
We first collect 25K images of portrait images which mainly contain self-occluded and soft shadows with minimum external occlusion on the body. 
%
%
We then apply our intermediate shadow removal model, which learns only from lightstage and synthetic human data, to these 25k images and then leverage these outputs as the pseudo ground truth for shadow-free images. 
We observed that such intermediate models already perform robustly for the portrait images with soft and self-occluded shadows.
By adding a novel shadow synthesized with 3D point lighting simulation (similar to the process used in the synthetic humans) onto the original input images, we construct the noisy portrait images with synthetic shadow and highlight as input, as shown in Figure~\ref{Fig:data}-(fourth) 
During the shadow simulation, we use the geometry information from monocular depth and surface normal detection modules, \textit{e.g.,}~\cite{ranftl2020towards}.

%% file: sec/4_experiments.tex
\section{EXPERIMENTS}
We perform extensive quantitative and qualitative comparisons of our method with existing portrait shadow removal methods and ablation studies for each module in our model. 

{\subsection{Implementation Details.}
Our model is implemented in PyTorch using 8x40GB(A100) at 512x512 resolution with 96 batch sizes in three steps. First, we initialize the Unet-based latent diffusion model, \textit{e.g.,} InstructPix2pix~\cite{brooks2023instructpix2pix}, and fine-tune the model using 400k data for light-aware harmonization captured from lightstage with lr=5e-5. Second, we fine-tune the model using 450k data for shadow removal created from lightstage (250k) and synthetic humans (200k) with lr=1e-5. Third, we further add 25k data from real-world scenes and fine-tune the model with lr=5e-6. The training takes 72h, 72h, and 24h for each step. We apply spatial data augmentation including rotation, flip, cropping, and padding for all steps. In inference time, we apply 50 diffusion steps. For portrait-segmentation, we use a transformer based model~\cite{wang2021pyramid} trained with internal data of image and foreground annotation. For lighting conditioning $\mathbf{G}$ in Figure~\ref{Fig:overview}-(b,c), we perform area-based downsampling of background image and input portrait image, respectively, to a 32x32.
}

{
\renewcommand{\tabcolsep}{2.2pt} 
\begin{table}[t]
\caption{\textbf{Comparison} with other methods on the validation data. \textbf{Bold} for the best, and \underline{underline} for the second best.}
\setlength{\extrarowheight}{2pt}
\centering
\begin{tabular}{|l||c|c|c|c|c|c|c|}
\hline
&GridNet &UNet &ResNet& ResNet$_p$ &HRNet &TFNet & Ours\\
\hline
\hline
\small{SSIM} $\uparrow$ & 0.828  & \underline{0.841} & {0.838} & {0.784}& 0.829&{0.834}& \textbf{0.883}\\
\hline
\small{LPIPS} $\downarrow$& 0.156&0.143 & {0.150} & {0.184}& 0.156 &\underline{0.137}&\textbf{0.093}\\
\hline
\end{tabular}
\label{table:comparison}
\end{table}
}

\subsection{Dataset.} We construct many validation and testing sets.
For validating our and other models with full ground truth, we collect the data from the lightstage and synthetic humans. We capture OLAT images of many new subjects and render the portrait images under a novel lighting distribution using unseen panorama environment maps whose corresponding shadow-free portrait images are rendered using the diffused panorama environment maps as described in Section~\ref{sec:data}. This mainly includes the portrait shadow by self-occlusion.
To validate the robustness of the model to the external shadows, we newly create a synthetic data using graphics simulations where we used unseen subjects and masks to render the portrait images under novel external shadows.
For testing, we collect many real-world portrait scenes from existing license-free stock data.
We also test our models on the real-world data collected from existing works \cite{zhang2020portrait,han2021deep}

\subsection{Metrics.} We use two metrics to measure the robustness of the shadow removal. LPIPS~\cite{zhang2018perceptual} measures the perceptual similarity between the ground-truth shadow-free images and the predictions, which measures the global shadow distributions. SSIM~\cite{wang2004image} scores the structure similarity between the ground truth and the prediction which emphasizes the local properties of images such as color and high-frequency details. Since our focus is on the foreground, we composite the prediction with the ground-truth background before measuring the scores.

{Other than validating the general shadow removal performance, we further evaluate the ability for the identity preservation and shadow removal consistency using experimental setups. To validate the identity preservation performance, we measure the cosine similarity of Adaface embedding features~\cite{kim2022adaface} between the deshadowed images and the ground truth. To validate the shadow removal consistency,  we create synthetic testing data where the three constant human models and background scenes are rendered under ten different shadows with diverse degrees and shapes, and we measure the shadow removal consistency based on the standard deviation of the LPIPS and SSIM metrics over different shadows as shown in Taable.}

\subsection{Comparative Evaluation}
\subsubsection{Baselines.} Since highly limited open sources are available for portrait shadow removal, we implement the baseline methods using public neural network architectures which are also widely adopted by many previous shadow removal works. UNet is used by the works from \cite{weir2022deep,futschik2023controllable,kim2024switchlight,nestmeyer2020learning,pandey2021total}, GridNet by \cite{zhang2020portrait}, ResNet by \cite{liu2022blind}, HRNet by \cite{weir2023lighting,ji2022geometry} and TransformerNet (TFNet) by \cite{chang2023tsrformer,guo2023shadowformer,hou2024compose}. {For ResNet, we follow the framework of gray scale shadow removal and colorization to fully reproduce the method from~\cite{liu2022blind}. We also report the performance of the ResNet directly pre-trained by the authors of~\cite{liu2022blind} using their dataset, which is denoted as ResNet$_p$.}
We train each model using the same training datasets as ours by minimizing the L1, VGG~\cite{gatys2016image}, and GAN~\cite{isola2017image} losses, which are widely used by the shadow removal and image processing works.

{
\renewcommand{\tabcolsep}{2.7pt} 
\begin{table}[t]
\caption{{\textbf{Identity preservation score comparison} with Adaface (AF) embedding cosine similarity metrics. \textit{w/o Up} means \textit{Ours} without guided upsampling. \textbf{Bold} for the best, and \underline{underline} for the second best.}}
\setlength{\extrarowheight}{2pt}
\centering
\begin{tabular}{|l||c|c|c|c|c||c|c|}
\hline
&GridNet &UNet &ResNet&HRNet &TFNet & Ours& w/o Up\\
\hline
\hline
AF $\uparrow$ & 0.699  & {0.715} & {0.713} & 0.728&{0.754}& \textbf{0.790}&\underline{0.785}\\
\hline
\end{tabular}
\label{table:identity_score}
\end{table}
}

For qualitative comparisons, we compare our methods with the model from Portrait Shadow Manipulation \cite{zhang2020portrait} and Deep Portrait Light Enhancement~\cite{han2021deep} using their respective precomputed results on the public dataset. 
We also perform the comparison of our methods with the model from Deep Portrait Delighting \cite{weir2022deep} and Controllable Light Diffusion \cite{futschik2023controllable} where respective authors provided the prediction results of their models on our testing data.

\subsubsection{Results.}
Table~\ref{table:comparison} summarizes the quantitative results among different methods. Please also refer to the supplementary for extensive results.
Our generative shadow removal model outperforms other methods by a large margin. Our method predicts the underlying appearance of shadow-free portraits in a globally coherent (low LPIPS) way, while effectively preserving the local appearance properties (high SSIM) such as details and colors.
In the qualitative comparison of Figure~\ref{Fig:validation}, we notice that other approaches suppress shadow and highlight well only when the input shadow is soft and the shadow boundary is smooth (e.g., 1\textsuperscript{st} row). However, when there exists a strong shadow or a novel external shadow on the portrait (e.g., 2\textsuperscript{nd} and 3\textsuperscript{rd} column), their results drop and leave visible remaining shadows.  
This behavior also presents in our real-world testing data where we show the visualization of the best two baselines, \textit{i.e.,} Unet and Transformer, in Figure~\ref{fig:qual} where they show significant weakness on strong shadows.
{The comparison between ResNet and ResNet$_p$ implies the importance of using our dataset where ResNet$_p$ demonstrates significant performance drops compared to ResNet due to the large domain gaps between the training and testing data.}

In Figure~\ref{Fig:comp_cecilia}, we provide the qualitative comparison with the method from \cite{zhang2020portrait} whose results are provided by the authors. 
Since GridNet (used in this work) is also designed to predict the local residual, it shows the weakness to the external and strong shadows. On the other hands, our method shows the better shadow removal results by generating globally coherent shadow-free portrait images. Interestingly, our model is highly generalizable to multiple people as shown in the right bottom of Figure~\ref{Fig:comp_cecilia} thanks to the nature of the large image prior. Figure~\ref{Fig:com_dpl} demonstrates the comparison with a light enhancement method \cite{han2021deep} where it marginally suppresses disturbing shadows and highlights, and our method shows much stronger shadow removal quality.

{
\renewcommand{\tabcolsep}{3pt} 

\begin{table}[t]
\caption{{\textbf{Shadow removal consistency comparison.} \textbf{Bold} for the best, and \underline{underline} for the second best.}}
\setlength{\extrarowheight}{2pt}
\centering
\begin{tabular}{|l||c|c|c|c|c|c|}
\hline
\textbf{Subject 1} &GridNet &UNet &ResNet &HRNet &TFNet & Ours\\
\hline
\hline
SSIM $\uparrow$ & 0.8184  & \underline{0.8389} & {0.8217}& 0.8255&{0.8189}& \textbf{0.8902}\\
Std ($\pm$) $\downarrow$& \underline{0.0447}  & {0.0465} & {0.0451}& 0.0496&{0.0535}& \textbf{0.0177}\\
\hline
LPIPS $\downarrow$& 0.1868&\underline{0.1659} &{0.1864}& 0.1809 &{0.1665}&\textbf{0.0844}\\
Std ($\pm$) $\downarrow$& \underline{0.0466}  & {0.0510} & {0.0470}& 0.0497&{0.0534}& \textbf{0.0187}\\
\hline
\end{tabular}

\begin{tabular}{|l||c|c|c|c|c|c|}
\hline
\textbf{Subject 2}&GridNet &UNet &ResNet &HRNet &TFNet & Ours\\
\hline
\hline
SSIM $\uparrow$ & 0.7091  & {0.7093} & \underline{0.7287}& 0.6968&{0.7027}& \textbf{0.7462}\\
Std ($\pm$) $\downarrow$& 0.0595  & \underline{0.0575} & {0.0641}& 0.0620&{0.0664}& \textbf{0.0321}\\
\hline
LPIPS $\downarrow$& 0.2090&0.2043 &\underline{0.1999}& 0.2136 &\underline{0.2014}&\textbf{0.1234}\\
Std ($\pm$) $\downarrow$& \underline{0.0415}  & {0.0442} & {0.0503}& 0.0479&{0.0473}& \textbf{0.0244}\\
\hline
\end{tabular}

\begin{tabular}{|l||c|c|c|c|c|c|}
\hline
\textbf{Subject 3}&GridNet &UNet &ResNet &HRNet &TFNet & Ours\\
\hline
\hline
SSIM $\uparrow$ & 0.7987  & \underline{0.8221} & {0.7983}& 0.7973&{0.7981}& \textbf{0.8780}\\
Std ($\pm$) $\downarrow$ & 0.0440  & {0.0451} & {0.0435}& 0.0438&\underline{0.0416}& \textbf{0.0240}\\
\hline
LPIPS $\downarrow$& 0.1979&\underline{0.1774} &{0.1964}& 0.2024 &{0.1874}&\textbf{0.0921}\\
Std ($\pm$) $\downarrow$ & 0.0481  & \underline{0.0482} & {0.0481}& \underline{0.0449}&{0.0467}& \textbf{0.0215}\\
\hline
\end{tabular}
\label{table:comparison}
\end{table}
}

We also demonstrate extensive comparisons with two state-of-the-art methods, Deep Portrait Delighting (DPD) \cite{weir2022deep} and Controllable Light Diffusion (CLD) \cite{futschik2023controllable}, on many real-world portrait images as shown in \ref{fig:qual} where respective authors provide the testing results.
DPD consistently underperforms, producing many artifacts such as saturation, blurs and whitening effects, and demonstrates many failure cases for the inputs with challenging shadow styles.
On the other hands, CLD performs well and highly generalizable to many scenes, performing best among other baseline methods. In particular, it could effectively remove diverse soft and self-occluded shadows. 
However, CLD suffers from the same problems as other local propagation methods for the case of strong and external shadows. For example, in Figure~\ref{fig:qual}-(a,b,c,d,g), the local propagation often falls into the local minima, leading to imperfect shadow removal; in Figure~\ref{fig:qual}-(c), the results are overly smooth, particularly for the regions with low intensity; in Figure~\ref{fig:qual}-(f), the lighting distribution of the foreground weakly matches that of background compared to our method. 
Our generative shadow removal framework is more effective in preserving the identity of the input portrait including skin color and details (\textit{e.g.,} Figure~\ref{fig:qual}-(e)), and shows strong appearance generation ability even under the colored shadow as shown in Figure~\ref{fig:qual}-(h).

{In Table~\ref{table:identity_score}, we summarize the identity preservation score for all the baseline methods. The results with our guided upsampling (\textit{Ours}) show the better identity score by restoring the original high-frequency details compared to the one without guided upsampling (\textit{w/o Up}); and our method consistently better preserves the identity compared to other baselines.
}

{In Table~\ref{table:comparison}, we also summarize the shadow removal consistency score of three testing subjects under diverse shadows for all baseline methods. This study demonstrates that the diffusion generated results (\textit{Ours} consistently outperforms other local propagation methods, and it is indeed robust to complex novel shadows by stably generating consistent shadow-free images based on the lowest standard deviation. We will also include visual results.
}

{\subsection{User Study}}
{To measure if our shadow removal model is indeed visual plausible from the human perception perspectives, we perform user study. To this end, we use 20 results sampled from Figure~\ref{fig:qual} and Figure.1-3 in the supplementary document. We use two state-of-the-art methods from  Deep Portrait Delighting ~\cite{weir2022deep} and Controllable Light Diffusion~\cite{futschik2023controllable} as baselines. We kindly note that Controllable Light Diffusion is proven to outperform all previous methods. 27 users participated to answer to three questions: Q1) Which result most effectively removes the shadows or highlights? Q2) Which result most effectively preserves the person’s identity (e.g., details and skin)? Q3) Which result best harmonizes with background scenes? 
}

{We summarize the percentage of how many times each method is selected in Table~\ref{table:user_study}. Overall, users select our method as the best result for all questions, implying that our method is perceptually effective to remove the portrait shadow and highlights; preserve the identity including the skin color and high-frequency details; and harmonize well with the background by preserving the original lighting distribution. We noticed that users sometimes perceive the portrait with some original shadows, \textit{i.e.,} the case when the shadow on the portrait is weakly removed, as more natural and better preserving the identity based on the selection percentage for the method from~\cite{futschik2023controllable}.
}

{
\renewcommand{\tabcolsep}{3pt} 
\begin{table}[t]
\caption{{\textbf{User study}: The summary of user study. \textbf{Bold} for the best, and \underline{underline} for the second best.}}
\setlength{\extrarowheight}{2pt}
\centering
\begin{tabular}{|l||c|c|c|c|}
\hline
&\cite{weir2022deep}& \cite{futschik2023controllable} & Ours\\
\hline
\hline
Q1 &1$\%$&\underline{16$\%$}   & \textbf{82$\%$}\\
\hline
Q2 &1$\%$&\underline{24$\%$}  &\textbf{75$\%$}\\
\hline
Q3 &1$\%$&\underline{24$\%$}  &\textbf{75$\%$}\\
\hline
\end{tabular}
\label{table:user_study}
\end{table}
}


\subsection{Ablation Studies}
We perform ablation studies to explore the importance of the modules in our compositional repurposing framework. Below, we describe the setup of each ablation along with results and analysis.

\subsubsection{Importance of Compositional Repurposing Framework.} We explore the importance of our compositional repurposing framework with several variants. 1) Joint learning (\textit{Joint}): we train the model using the background harmonization and portrait shadow removal data together. 2) Learning without harmonization (\textit{w/o Harmo}): we train the model only with our shadow-removal data without harmonization. 

In Table~\ref{table:ablation}, we notice that mixing the harmonization and shadow removal data largely drops the performance. The training of two different tasks at the same time results in suboptimal shadow removal quality as shown in Figure~\ref{abl:JR}.
\textit{w/o Harmo} shows meaningful gaps with our full model in terms of perceptual score (LPIPS). This means that training with harmonization data strengthens the modeling of global shadow distribution by learning the appearance of many portraits under different lighting conditions. As shown in Figure~\ref{abl:no_harmo}, compared to \textit{w/o Harmo}, our method is effective to model the globally coherent shadow-free appearance that well matches the background distribution.

{The decomposition into harmonization and shadow removal is important to effectively model the lighting harmonization and shadow removal. In particular, the harmonization helps with modeling of background lighting prior and correlating such prior to the intrinsic appearance of humans under minimum shadows and highlights. Based on our ablation studies, if the model jointly learns with two tasks, it is often biased to the harmonization task; and if the model only learns with shadow removal task, the visual statistics of the generated portraits are often biased to the distribution of the training data, leading to sub-optimal prediction results with significant identity distortion and degraded robustness to novel shadows.
}

{
\renewcommand{\tabcolsep}{3pt} 
\begin{table}[t]
\caption{\textbf{Ablation study}: the importance of our compositional repurposing frameworks and other important components. \textbf{Bold} for the best, and \underline{underline} for the second best.}
\setlength{\extrarowheight}{2pt}
\centering
\begin{tabular}{|l||c|c|c|c|c|}
\hline
&Joint& w/o Harmo &w/o Up & Ours\\
\hline
\hline
SSIM $\uparrow$  &0.837&\underline{0.880}   & 0.866& \textbf{0.883}\\
\hline
LPIPS $\downarrow$&0.136&0.0975 &\textbf{0.090}  &\underline{0.093}\\
\hline
\end{tabular}
\label{table:ablation}
\end{table}
}

\subsubsection{Contribution of Each Dataset.} Given our compositional repurposing framework (after the step of harmonization repurposing), we study how each shadow removal dataset contributes to the appearance modeling. 1) Learning from lightstage (\textit{only Light}): We train the model with only lightstage data with real humans. 2) Learning from synthetic humans (\textit{only Synth}): We only use synthetic humans and shadows for training. 3) Learning without real data (\textit{w/o Real}): We do not train our model with the data from many real humans.

Table~\ref{table:ablation2} summarizes the performance for each dataset ablation. From \textit{only Light}, the large gaps with ours in LPIPS score identify that the synthetic human datasets with perfect ground-truth pairs are indeed useful to develop the shadow removal model that is highly robust to diverse shadows. In Figure~\ref{abl:no_synth}, the model learned only from lightstage data often includes highlighting artifacts that mimic the one-point lighting in the lab environment.
However, it does not mean the lightstage data is not necessary. As visualized in Figure~\ref{abl:no_ligghtstage}, the generation results from the model that only uses synthetic humans often look fake, and it sometimes (though that that many) completely changes the color distribution for a specific body part (\textit{e.g.,} hairs).  
While the quantitative gains by learning from real data with pseudo ground truth is marginal in terms of SSIM (although there exists ignorable amount of performance drops in LPIPS), its qualitative gains are sometimes significant as shown in Figure~\ref{abl:no_real} where our model learned with real data could handle diverse input shadow styles and identities (\textit{e.g.,} skin colors).

\subsubsection{Importance of Guided Upsampling.} Without applying guided upsampling (\textit{w/o Up}): We use the results that are directly generated from the diffusion model without any post-processing. In Table~\ref{table:ablation}, the results with guided upsampling (Ours) performs the best in terms of SSIM. This means that the local details such as wrinkles and clothing textures as shown in Figure~\ref{abl:up} of the results with upsampling have better matches that one without upsampling. Such high-frequency details are indeed crucial for production-level applications since they are highly correlated to the identity. Our guided upsampling module, however, sometimes restores the shadow boundary as artifacts as shown in Figure~\ref{abl:up} when there exists imperfection on the shadow removal from our diffusion model. This weakens the perceptual quality of the shadow removal results in Table~\ref{table:ablation} where the direct generation from our diffusion model demonstrates the best shadow removal quality in terms of LPIPS. Please refer to the supplementary for more visual comparison between with and without upsampling.

\subsection{Application}
Our shadow removal model improves the quality of various downstream applications. In Figure~\ref{application1}, (left) we perform text-guided relighting \cite{iclight} using the prompt of \textit{neon light, city} where our shadow-free portrait images highly improve the visual relighting quality; (middle) we also demonstrate that our method can improve the accuracy of the portrait parsing results; (right) and enables the clean appearance modeling. Using the portrait parsing labels, we can also demonstrate a part-aware shadow editing in Figure~\ref{application2}.

{
\renewcommand{\tabcolsep}{3pt} 
\begin{table}[t]
\caption{\textbf{Ablation study}: the importance of each training dataset. \textbf{Bold} for the best, and \underline{underline} for the second best.}
\setlength{\extrarowheight}{2pt}
\centering
\begin{tabular}{|l||c|c|c|c|}
\hline
&   only Light&only Synth & w/o Real  & Ours\\
\hline
\hline
SSIM $\uparrow$ &  0.843 & 0.871 &\underline{0.882}  & \textbf{0.883}\\
\hline
LPIPS $\downarrow$&0.1327 & 0.1086& \textbf{0.0934}&\underline{0.0933}\\
\hline
\end{tabular}
\label{table:ablation2}
\end{table}
}

\subsection{Limitation}
In Figure~\ref{fig:limitation}, (a) our model shows a weakness in preserving the original skin color when processing an input portrait; (b) it is sometimes confused by the colored texture on the skin, \textit{e.g.,} cosmetic; and (c) there exist notable quality drops with the input portrait images with significant missing regions. {While our model reasonably handles multi-person (shown in Figure~\ref{Fig:comp_cecilia}) thanks to the pretrained nature of diffusion models, the full performance is not ensured due to the nature of the training data (\textit{i.e.,} portrait scenes). 
If there exists insufficient context of the background scene, \textit{e.g.,} the face is covering almost the entire image, our model borrows the lighting cues mainly from the foreground. Although guided refinement module highly effective to restore the high-frequency details of the person from the input image to preserve the identity, it sometimes restore the unwanted shadow as well as shown in Figure~\ref{abl:up}. Also, if no information is available due to very strong shadow or highlights, \textit{e.g.,} the pixel value is almost zero, our guided refinement module inherently cannot restore the original details due to the missing information from the input image as shown in Figure~\ref{fig:limitation}-(c) }

%% file: sec/5_conclusion.tex
\section{CONCLUSION}
In this paper, we introduce a shadow removal model that can remove the disturbing shadows and highlights in a globally coherent way.
We address the fundamentally ill-posed local shadow removal problem by casting this as a generative task where a diffusion model re-builds a shadow-free image from noises by referring to the input portrait image.
To develop our generative model that can robustly remove diverse shadows while maintaining original lighting conditions from the input image, we perform a series of training using background harmonization and shadow removal data.
To enable such compositional repurposing framework, we construct diverse training dataset from a lightstage system and graphics simulation.
As a post-processing, our guided upsampling module restores the original high-frequency details.
We perform extensive comparisons and ablative studies where our model demonstrates strong robustness, effective identity preservation, naturalness, and generalization.

%% file: sec/6_supple.tex
\begin{figure}
    \centering
    \includegraphics[width=1\columnwidth]{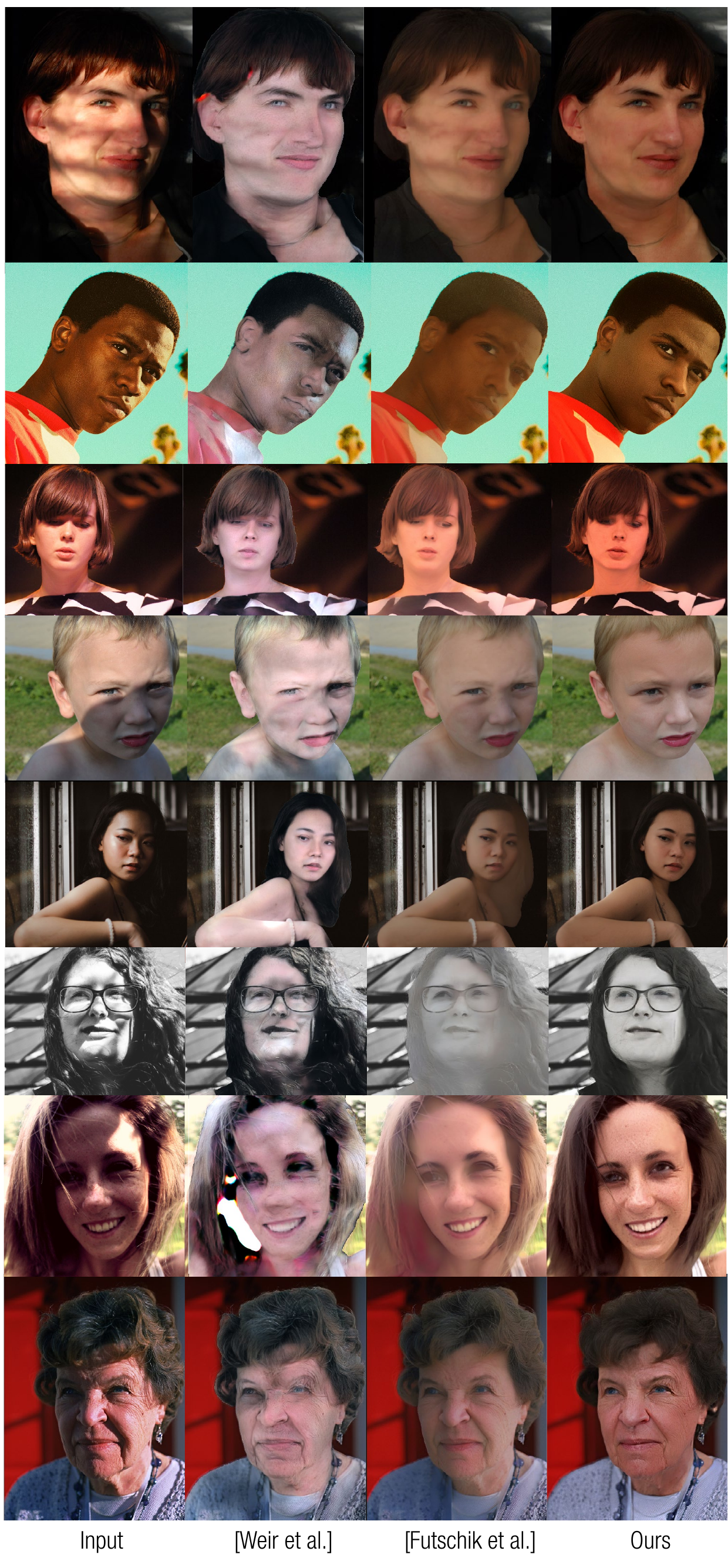}
    \vspace{-6mm}
    \caption{\textbf{Qualitative comparison} with Deep Portrait Delighting \cite{weir2022deep} and Controllable Light Diffusion \cite{futschik2023controllable}. }
    \label{fig:more_res}
\end{figure}

\begin{figure}[t]
    \centering
    \includegraphics[width=0.9\columnwidth]{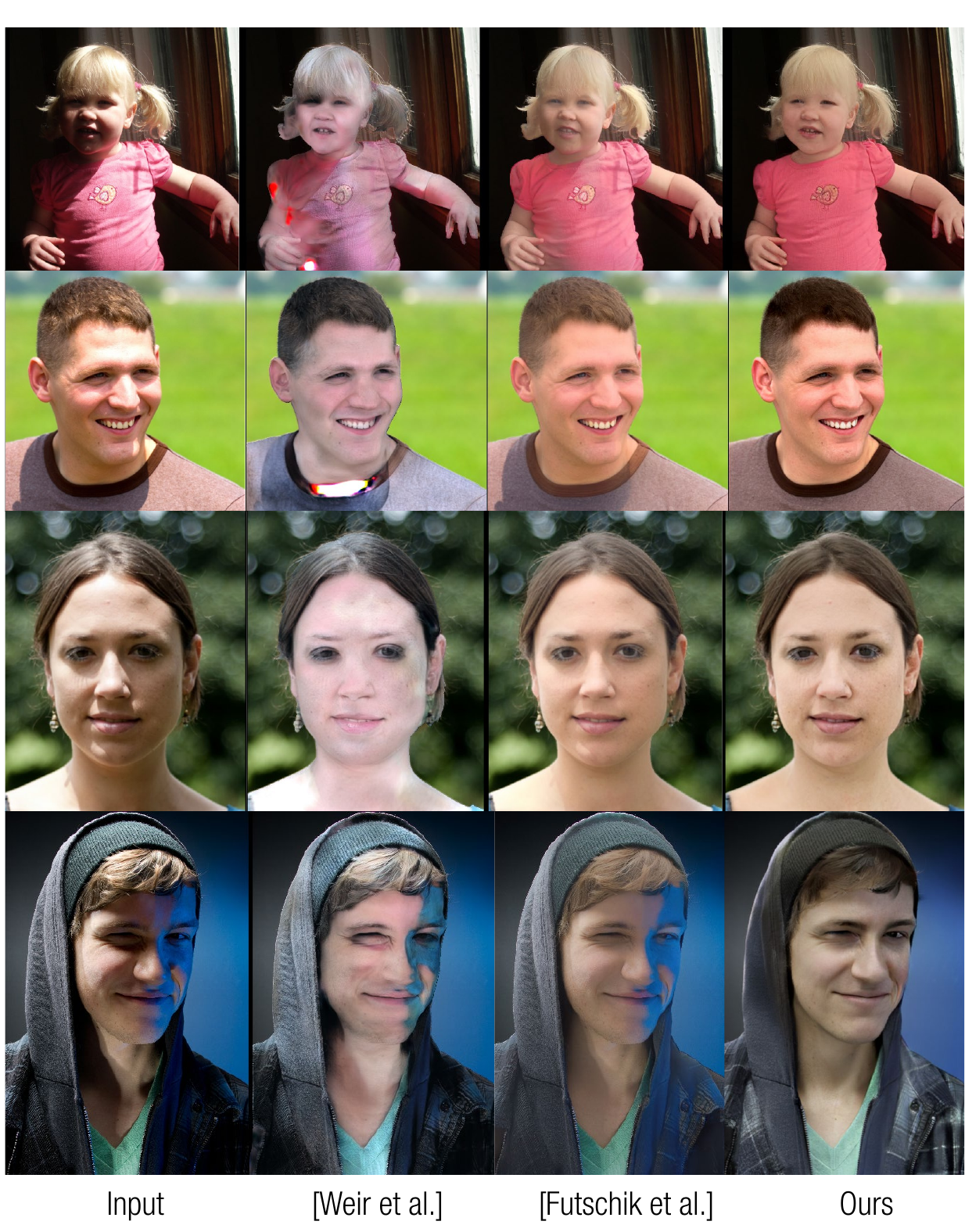}
    \vspace{-4mm}
    \caption{\textbf{Qualitative comparison} with Deep Portrait Delighting 
    \cite{weir2022deep} and Controllable Light Diffusion \cite{futschik2023controllable}. }
    \label{fig:more_res2}
\end{figure}

\begin{figure*}[t]
    \centering
    \includegraphics[width=1.75\columnwidth]{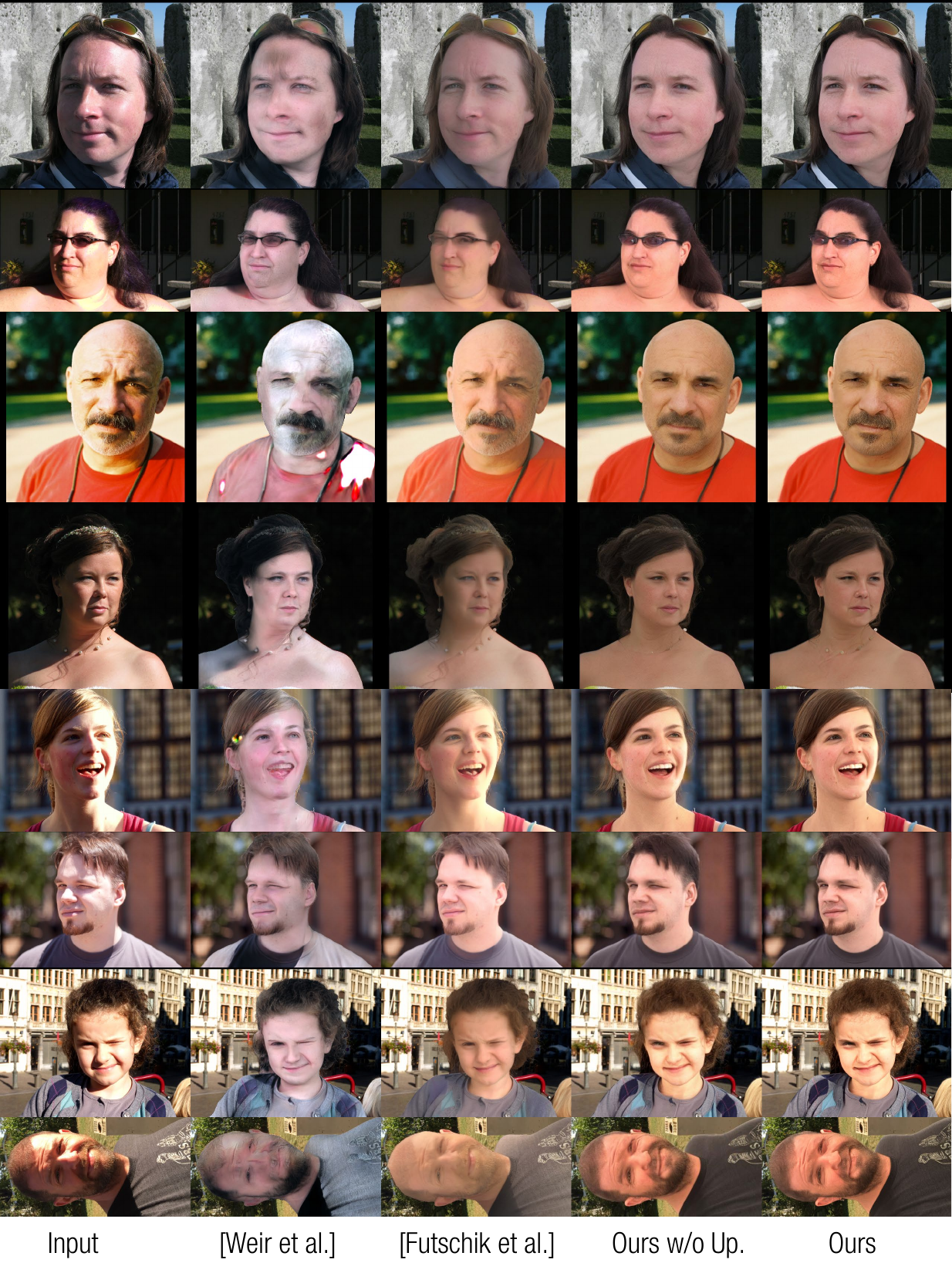}
    \caption{\textbf{Comparison} with Deep Portrait Delighting \cite{weir2022deep}, Controllable Light Diffusion \cite{futschik2023controllable}, and ours without guided upsampling. }
    \label{fig:more_res3}
\end{figure*}

\begin{figure*}[t]
    \centering
    \includegraphics[width=1.75\columnwidth]{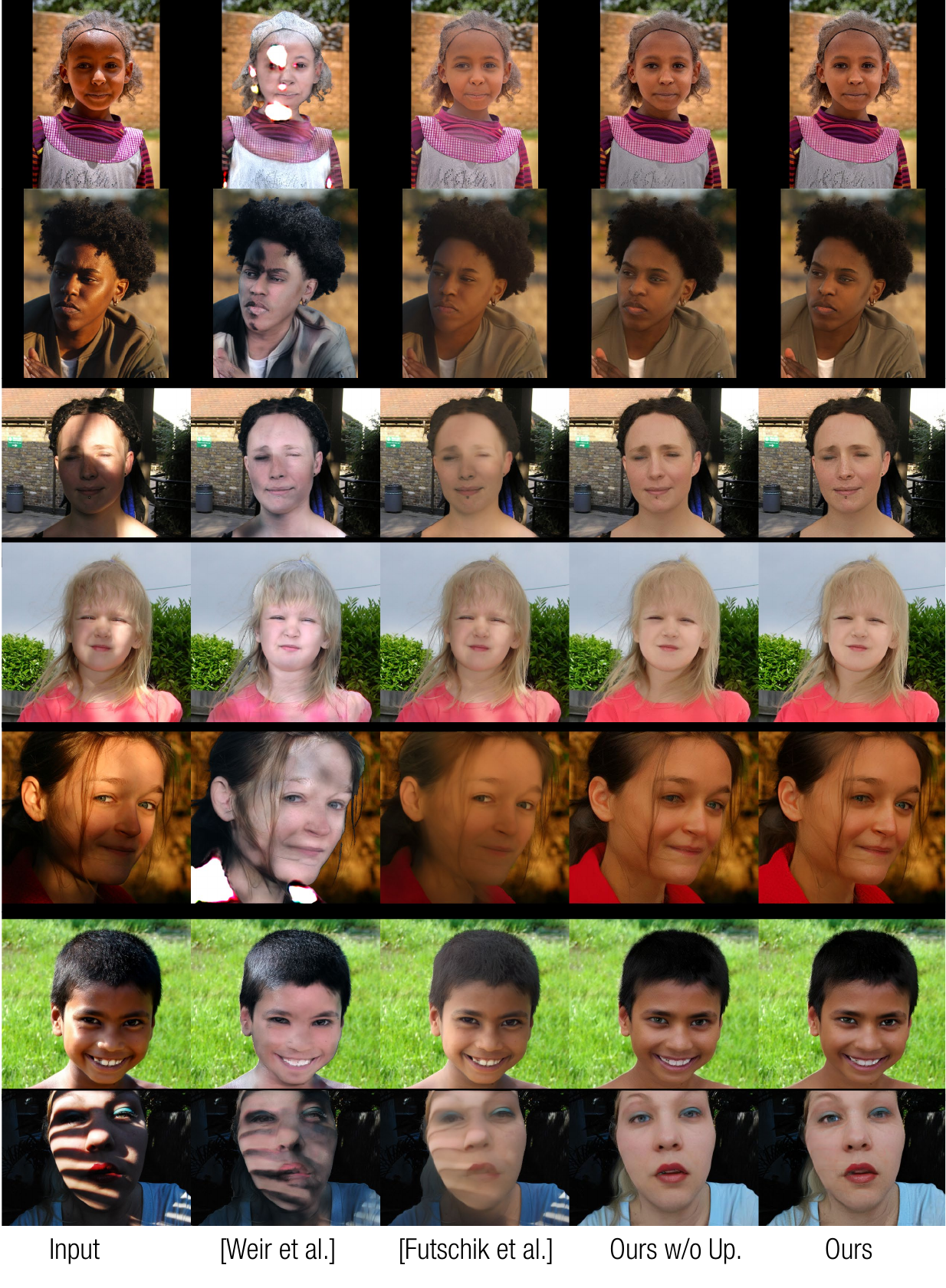}
    \vspace{-5mm}
    \caption{\textbf{Comparison} with Deep Portrait Delighting
    \cite{weir2022deep}, Controllable Light Diffusion \cite{futschik2023controllable}, and ours without guided upsampling. }
    \label{fig:more_res4}
\end{figure*}

\begin{figure*}[t]
    \centering
    \includegraphics[width=1.75\columnwidth]{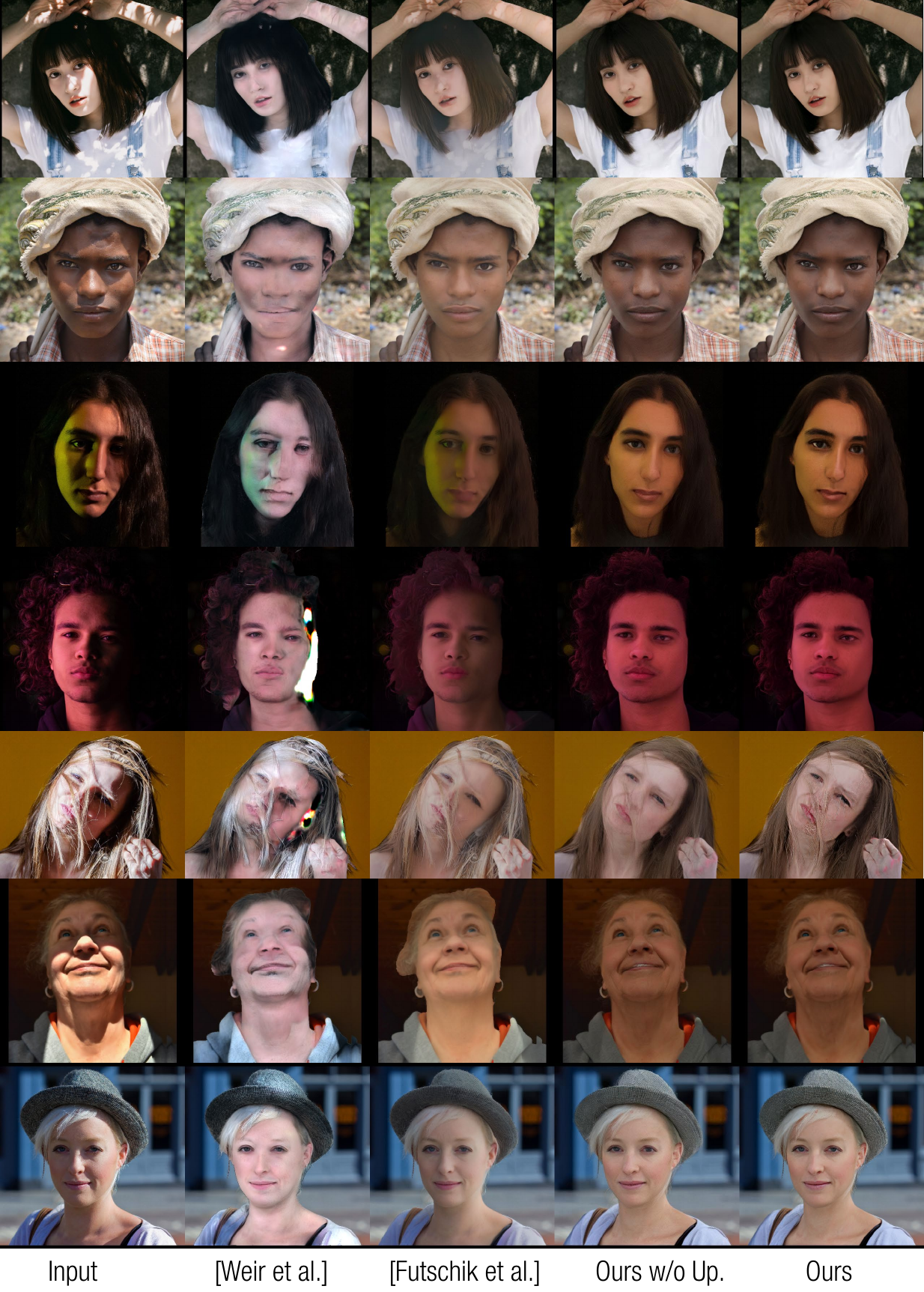}
    \vspace{-3mm}
    \caption{\textbf{Comparison} with Deep Portrait Delighting
    \cite{weir2022deep}, Controllable Light Diffusion \cite{futschik2023controllable}, and ours without guided upsampling. }
    \label{fig:more_res5}
\end{figure*}